\pdfoutput=1 
\RequirePackage{fix-cm}
\documentclass[natbib,twocolumn]{svjour3}          %
\smartqed  %
\usepackage{graphicx}
\usepackage{xspace}
\usepackage{multirow}
\usepackage{rotating}
\usepackage{subfigure}
\usepackage{amsmath}
\usepackage{url}
\usepackage{amssymb}
\usepackage{epsfig}
\usepackage{rotating} %
\usepackage{color}
\usepackage{booktabs} %
 
\graphicspath{{./}}
\usepackage[english,american]{babel}

\newcommand{\midruleMPI}{\cmidrule(lr){1-1}  \cmidrule(lr){2-7} \cmidrule(lr){8-8}}

\newcommand{\myparagraph}[1]{ \paragraph{#1}} %

\newcommand{\MpiDs}{MPII Cooking\xspace}
\newcommand{\EccvDs}{MPII Composites\xspace}
\newcommand{\MpiNew}{MPII Cooking~2\xspace}
\newcommand{\Scriptknowledge}{Script data\xspace}
\newcommand{\tfidf}{{tf$*$idf}\xspace}

\newcommand{\DBnFrames}{2,881,616\xspace}

\newcommand{\DBhours}{27\xspace}
\newcommand{\DBnSubjects}{30\xspace} %
\newcommand{\DBnVideoSeq}{273\xspace}  %
\newcommand{\DBnActivities}{67\xspace}
\newcommand{\DBnObjects}{155\xspace}
\newcommand{\DBnAttributes}{222\xspace} %
\newcommand{\DBnTasks}{59\xspace} %
\newcommand{\DBnGtInterval}{14,105\xspace} %
\newcommand{\DBnAttributeInstances}{54,774\xspace} %

\newcommand{\ApproachName}[1]{Propagated Semantic Transfer}

 \makeatletter
 \DeclareRobustCommand\onedot{\futurelet\@let@token\@onedot}
 \def\@onedot{\ifx\@let@token.\else.\null\fi\xspace}
 \def\eg{e.g\onedot} 
 \def\ie{i.e\onedot}

 \def\etal{\textit{et~al\onedot}}

 \makeatother

\DeclareRobustCommand{\figref}[1]{Figure~\ref{#1}}
\DeclareRobustCommand{\figsref}[1]{Figures~\ref{#1}}

\DeclareRobustCommand{\Figsref}[1]{Figures~\ref{#1}}

\DeclareRobustCommand{\secref}[1]{Section~\ref{#1}}

\DeclareRobustCommand{\secsref}[1]{Sections~\ref{#1}}

\DeclareRobustCommand{\tableref}[1]{Table~\ref{#1}}

\DeclareRobustCommand{\tablesref}[1]{Tables~\ref{#1}}

\DeclareRobustCommand{\eqnref}[1]{Equation~(\ref{#1})}

\DeclareRobustCommand{\eqnsref}[1]{Equations~(\ref{#1})}

\journalname{IJCV}
\begin{document}

\title{Recognizing Fine-Grained and Composite Activities\\ using Hand-Centric Features and Script Data}

\subtitle{}

\author{Marcus Rohrbach  \and Anna Rohrbach \and Michaela Regneri \and Sikandar~Amin \and Mykhaylo Andriluka \and Manfred Pinkal \and Bernt Schiele        
}

\institute{Marcus Rohrbach\textsuperscript{1,2} \and  Anna Rohrbach\textsuperscript{2} \and  Michaela Regneri\textsuperscript{3,6} \and Sikandar~Amin\textsuperscript{2,4} \and  Mykhaylo Andriluka\textsuperscript{2,5} \and  Manfred Pinkal\textsuperscript{3} \and  Bernt Schiele\textsuperscript{2}\at
\textsuperscript{1} UC Berkeley EECS and ICSI, Berkeley, CA, United States.\\
            \textsuperscript{2} Max Planck Institute for Informatics, Saarbr\"ucken, Germany.\\
             \textsuperscript{3} Saarland University, Department of Computational Linguistics and Phonetics, Saarbr\"ucken, Germany.\\
           \textsuperscript{4} Technische Universit\"at M\"unchen, Department of Informatics, Germany.\\
           \textsuperscript{5} Stanford University, CA, United States.\\
          \textsuperscript{6} SPIEGEL-Verlag, IT Department, Hamburg, Germany, this research was carried out at Saarland University.\\
}

\date{\ }%

\maketitle

\begin{abstract}
Activity recognition has shown impressive progress in recent years. However, the challenges of detecting fine-grained activities and understanding how they are combined into composite activities have been largely overlooked. In this work we approach both tasks and present a dataset which provides detailed annotations to address them. The first challenge is to detect fine-grained activities, which are defined by low inter-class variability and are typically characterized by fine-grained body motions. We explore how human pose and hands can help to approach this challenge by comparing two pose-based and two hand-centric features with state-of-the-art holistic features. To attack the second challenge, recognizing composite activities, we leverage the fact that these activities are compositional and that the essential components of the activities can be obtained from textual descriptions or scripts. 

We show the benefits of our hand-centric approach for fine-grained activity classification and detection. For composite activity recognition we find that decomposition into attributes allows sharing information across composites and is essential to attack this hard task. Using script data we can recognize novel composites without having training data for them. 
\end{abstract}

\section{Introduction}
\label{sec:intro}

\newlength{\trimteaserx}
\setlength{\trimteaserx}{0pt}

\newlength{\trimteasery}
\setlength{\trimteasery}{2cm}
\begin{figure*}[t]
\begin{center}
  \includegraphics[trim=\trimteaserx{} 1.4\trimteasery{} \trimteaserx{} 1.3\trimteasery{}, clip=true, width=0.9\textwidth]{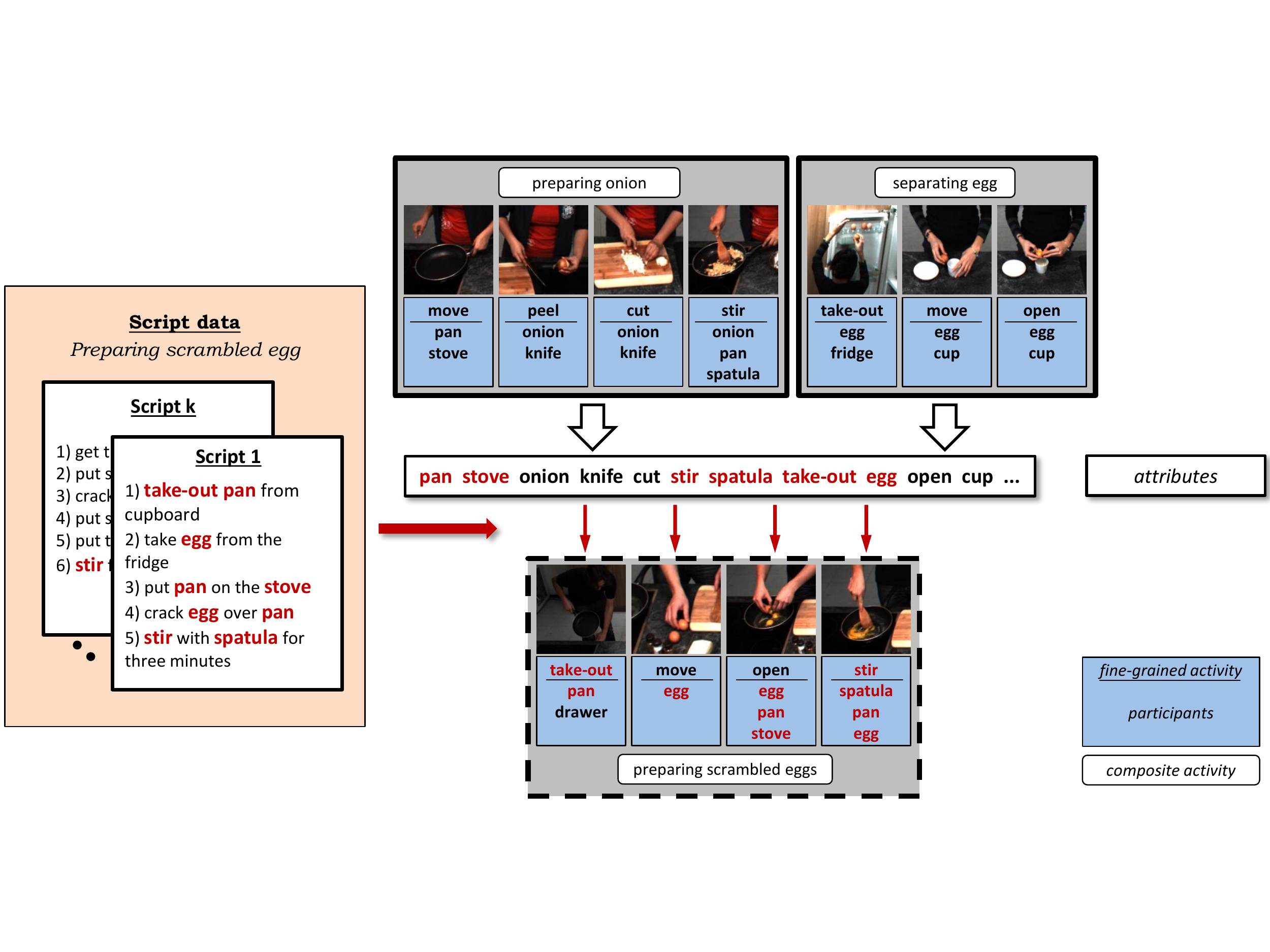}
  \caption[Sharing or transferring attributes of composite activities using script data.]{Sharing or transferring attributes of composite activities using script data. Composite activities (gray boxes) are composed of activities and their participants (light-blue boxes), modeled as attributes. These attributes can be transferred to unseen composite activities (dashed-line box) with the help of script data which allows estimating the relevant attributes (red). Our activities have the additional challenge of being fine-grained, we thus refer to them as fine-grained activities.}
  \label{fig:teaser}
\end{center}
\end{figure*}

Human activity recognition in video is a fundamental problem in computer vision.
State-of-the-art methods
\citep[\eg][]{tang12cvpr,wangYuTang13iccv,wang13iccv,karpathy14cvpr} achieve near perfect results for simple actions \citep[\eg KTH dataset,][]{schuldt04icpr} and robustly recognize actions in realistic settings
such as Hollywood movies \citep{marszalek09cvpr}, videos from YouTube \citep{liu09cvpr}, or sport
scenes \citep{rodriguez08cvpr}.

While impressive progress has been made, we argue that 
most works are addressing only a part of the overall activity recognition challenge. Many application scenarios, such as human-robot interaction or elderly care require to understand complex activities (\eg \emph{does the person prepare food?}), consisting of multiple fine-grained activities and object manipulations (\eg \emph{is it fried and what is in it?}). Frequently it is important to recognize both, the individual steps and the high level composite activities, \eg as we have shown for the task of video description \citep{rohrbach14gcpr}. 
Consequently we approach both problems in this work: recognizing fine-grained activities and recognizing composite activities. \emph{Fine-grained activities} are defined as a set of activities which are visually very similar, \ie have a low inter-class variability. \emph{Composite activities} are activities which can be temporally decomposed into multiple shorter activities, \ie they consist of multiple steps.  
We note that both the terms are not exclusive, \ie composite activities can also be fine-grained. In fact some of our composites are very similar. However, in our work we consider composite activities which consist of fine-grained activities.

When surveying the field we also noticed a lack of datasets allowing to pursue the challenges of fine-grained and composite activity recognition. Specifically this is reflected in the following limiting  
factors of current benchmark databases. 
First, while datasets with large numbers of activities exist, the typical inter-class variability is high. This seems rather unrealistic for many domains such as surveillance or elderly care where we need to differentiate between consequentially different but visually similar activities \eg \emph{hug someone} versus \emph{hold someone} or  \emph{throw in garbage} versus \emph{put in drawer}. Second, the activities considered so far are full-body activities, \eg \emph{jumping} or \emph{running}. This appears rather untypical for many applications where we want to differentiate between more small motion  and frequently hand centric activities. Consider \eg the \emph{cutting activity} in domains such \emph{cooking}  (see \figref{fig:teaser}), \emph{handicraft work} or \emph{surgeries}, as well as different \emph{repairing} activities in the domain of \emph{house keeping} or  \emph{machine maintenance} with subtle difference in motion and low inter-class variability. 
As a third limitation we found that many available databases contain videos of few second length %
and focus on simple basic-level activities such as \emph{walking} or \emph{drinking}. In contrast, the recognition of longer-term, complex, and composite activities such as \emph{assembling furniture}, \emph{food preparation}, or \emph{surgeries} have been rarely addressed in computer vision.
Notable exceptions exist (see \secref{sec:related})
even though these have other limiting factors such as small number of classes.

In this work, which is an extension of our original publications \citep{rohrbach12cvpr} and \citep{rohrbach12eccv}, we recorded, annotated, and publicly released a large-scale dataset in a kitchen scenario which addresses the discussed limitations. This allows us to work on the challenges of fine-grained and composite activity recognition as follows.

Recognizing fine-grained activities is challenging due to their low inter-class variability. In contrast to fine-grained object recognition challenges where the same object category typically is also visually consistent, activities of the same category are frequently very diverse, \ie have a high intra-class variability. Consider \eg the activities \emph{peeling}, which can be very different depending of the participating object: \emph{peeling a carrot} versus \emph{peeling a pineapple}. 
At the same time, we have to handle small differences between categories, \ie low inter-class variability, consider \eg \emph{mix} versus \emph{stir} or \emph{slice} versus \emph{cut dice}.
This typically requires to understand the difference between fine-grained body motions.
To approach both of these challenges we propose to focus on body pose and hands. 
As can be seen in \figsref{fig:teaser} and~\ref{fig:dataset} many fine-grained activities, especially in our kitchen scenario, are hand-centric. Here it is not only important to understand the activity but also the participating object, \eg  \emph{open egg} versus \emph{open tin}. We thus propose to focus on the hand regions for extracting visual features. However, hand detection is a challenging problem in itself in real-world scenarios due to a large variability in shape and frequent partial occlusions 
\citep{mittal11bmvc,gkioxari13cvpr}. 
To get reliable hand detections, we 
integrate a hand detector into an articulated pose estimation.
Consequently we use the hand position to extract color Sift and Dense Trajectories \citep{wang13ijcv} and learn detectors for fine-grained activities and their participating objects.
Recently, \citet{jhuang13iccv} showed that exploiting body pose in form of body joints can be beneficial for full-body activities.
We explore two approaches based on body pose tracks, motivated from work in the sensor-based activity recognition community~\citep{zinnen09iswc}. 

For recognizing composite activities, state-of-the-art methods, which build on discriminative learning from low-level activity features, experience scalability issues due to the typically highly diverse composite activities and little training data.
A promising approach towards scaling activity recognition methods to a large number of complex activities is to use intermediate representations that are shared and transferred across activities by exploiting their compositional nature. We exploit this technique and propose building on
an attribute-based representation, with attributes denoting the fine-grained activities and the participating objects. For example in \figref{fig:teaser} the composite activity \emph{preparing scrambled egg} shares the attributes \emph{stir} and \emph{spatula} with the composite activity  \emph{preparing onion} and the attributes \emph{open} and \emph{egg} with the composite activity  \emph{separating egg}. Instead of learning a holistic model for each composite activity we learn
models for a large set of attributes shared across composite activity classes. Such approaches have
been shown effective to recognize previously unseen object categories \citep{lampert13pami} 
and have also been applied to activity recognition \citep{liu11cvpr}. 
A major challenge to recognize everyday activities is that these
composite activities can often be performed in a wide variety of ways, 
and it is practically infeasible to create a visually annotated training set with all possible alternatives.
Instead, we %
collect a large number of textual descriptions (scripts) for a composite activity to compute the association strength between attributes and composite activities. Using this script data we can not only handle the inherent variation of composites but also recognize unseen composite activities. As illustrated in \figref{fig:teaser}, the attributes in red are determined to be important for  \emph{preparing scrambled eggs} using script data and can be transferred from known composites such as \emph{separating egg} and \emph{preparing onion}.

Our main contributions are as follows. First, we propose several hand- and pose-based activity recognition approaches to recognize fine-grained activities and their object participants. We benchmark them together with state-of-the-art activity recognition features on our dataset. 
Second, we contribute an attribute-based approach which shares knowledge across composite activities and exploits textual script data to handle their large variability and allows transfer to unseen composite activities. 
Third, we recorded and annotated a video dataset called \emph{\MpiNew}. It provides challenges for classification and detection of fine-grained activities and their participants, human pose estimation, and composite activity recognition (optionally) using script data. In addition to activity recognition, which is the focus of this work, the dataset is also being used for 3D human pose estimation \citep{amin13bmvc}, multi-frame pose estimation \citep{cherian14cvpr}, discovering object categories from activities \citep{srikantha14eccv}, grounding semantic similarities of natural language sentences in video \citep{regneri13tacl}, and for generating natural language descriptions \citep{rohrbach13iccv,rohrbach14gcpr}.

The remaining article is structured as follows. We first make an extensive review of related datasets, activity recognition approaches, and the use of text data for visual recognition in \secref{sec:related}. Then we introduce our \MpiNew dataset in \secref{sec:dataset} which we benchmark in the 	subsequent sections. In \secref{sec:approach:handAndpose} we make a quantitative comparison of our pose-recognition and hand detection with related work on the pose challenge of our dataset. Using the pose-estimation and hand detections  we define  several visual features and discuss fine-grained activity detection in \secref{sec:approach:finegrained}. 
In \secref{sec:approach} we present our approach to combine the fine-grained activities to composite activities and integrate script data. In \secref{sec:eval} we evaluate fine-grained and composite activity recognition and then we conclude with the most important findings and directions for future work in \secref{sec:conlusion}.

\section{Related work}
\label{sec:related}
We first present an overview of the different video activity recognition datasets (\secref{sec:cvpr12:related:datasets}) and then review recent approaches to activity recognition (\secref{sec:related:activityRecognition}), putting a focus on works which use human pose as a cue. Next we discuss works which use textual information for improved recognition of activities (\secref{sec:related:textForActivity}). We conclude by relating them to our work (\secref{sec:related:dscripts:relations}).

\newcommand{\datasetmidrule}{\cmidrule(){1-9}} 

\begin{table*}
\center
\begin{small}
\hspace{-2mm}
\begin{tabular}{l r@{}l r r @{}l r r c}
\toprule
 Dataset& cls&,det&       classes &clips&/videos&\multicolumn{1}{@{}c@{}}{subjects}&  \multicolumn{1}{@{}c@{}}{\# frames}& \multicolumn{1}{@{}c@{}}{resolution}\\    
\datasetmidrule
\multicolumn{5}{@{\ }l}{\textbf{Full body pose datasets}}	\\
KTH~\citep{schuldt04icpr} & cls&& 6 & 2,391&  &  25  & $\approx$200,000 &     160x120 \\
USC gestures~\citep{natarajan08cvpr} &cls&&6 & 400& & 4 & &740x480\\
MSR action~\citep{yuan09cvpr}& cls&,det & 3 &63&  & 10 &  & 320x240\\ 
\datasetmidrule
\multicolumn{5}{@{\ }l}{\textbf{Movie and web video datasets}}\\
Hollywood2~\citep{marszalek09cvpr}& cls&& 12 & 1,707&/69 &   \\
UCF 101 \citep{soomro12arxiv} & cls&& 101 & 13,320& & 	& $\approx$2,400,000& 320x240\\
Sports-1M~\citep{karpathy14cvpr} & cls & &487 & 1.1 mil&\\
HMDB51~\citep{kuehne11iccv}& cls&& 51 & 6,766& &  &  & height:240\\
ASLAN~\citep{kliper12pami}& cls&&  432 & 3,631&/1,571 & &\\ 
Coffee and Cigarettes~\citep{laptev07iccv}& &\ det & 2 & 264&/11 &  &  \\
High Five~\citep{patron10bmvc} & cls&,det & 4 & 300&/23 & & \\
MPII Movie Description~\citep{rohrbach15cvpr} & cls &,det &  &68,327&/94 & & & 1920x1080 \\
\datasetmidrule \multicolumn{5}{@{\ }l}{\textbf{Surveillance datasets}}\\
PETS 2007~\citep{ferryman07pets}& &\  det & 3 &  10 &  & &  32,107 & 768x576 \\ %
UT interaction~\citep{ryoo09iccv}& cls&,det & 6 & 120& & 6   \\
VIRAT~\citep{oh11cvpr} & &\   det & 23 & 17& & &1920x1080\\
\datasetmidrule \multicolumn{5}{@{\ }l}{\textbf{Assisted daily living datasets}}\\
TUM Kitchen~\citep{tenorth09iccw} & &\ det & 10 & 20  &  /4  & &36,666 &   384x288 \\ %
CMU-MMAC~\citep{torre09tr} & cls&,det &  $>$130& & ~26  & && 1024x768~~\\
URADL~\citep{messing09iccv} & cls&& 17 & 150&/30 &  5  & $\leq$	50,000 &   1280x720~~  \\
\MpiNew (our dataset) & cls&,det & \DBnActivities / \DBnTasks  & \DBnGtInterval &/\DBnVideoSeq &  \DBnSubjects   & \DBnFrames &  1624x1224 \\ %
\bottomrule  \\
\end{tabular} 
\end{small}  
\caption[Overview of activity recognition datasets]{Overview of activity recognition datasets: We list if datasets allow for classification (cls), detection (det); number of activity classes; number of clips extracted from full videos (only one listed if identical), number of subjects, total number of frames, and resolution of videos. We leave fields blank if unknown or not applicable.} 
\label{tbl:cvpr12:datasets} 
\end{table*} 

\subsection{Activity Datasets} 
\label{sec:cvpr12:related:datasets}
Even when excluding single image action datasets such as the Stanford-40 Action Dataset \citep{yao11iccv} or the Pascal Action Classification Challenge \citep{everingham11pascal}, the number of proposed activity datasets is quite large (\citet{chaquet13cviu} survey 68 datasets). Here, we 
focus on the most important ones with respect to database size, usage, and similarity to our proposed dataset (see \tableref{tbl:cvpr12:datasets}). 
We distinguish four broad categories of datasets: full body pose, movie and web, surveillance, and assisted daily living datasets -- our dataset falls in the last category.

The full body pose datasets are defined by actors performing full body actions. 
KTH~\citep{schuldt04icpr}, USC gestures~\citep{natarajan08cvpr}, and similar datasets~\citep{singh11iccv} require classifying simple full body and mainly repetitive activities. 
 The MSR actions~\citep{yuan09cvpr} pose a detection challenge limited to three classes.
In contrast to these full body pose datasets, our dataset contains more and in particular fine-grained activities.

The second category consists of movie clips or web videos with challenges such as partial occlusions, camera motion, and diverse subjects. UCF50\footnote{\label{fn:ucf50}http://vision.eecs.ucf.edu/data.html} and similar datasets~\citep{liu09cvpr,niebles10eccv,rodriguez08cvpr} focus on sport activities. Kuehne \etal's evaluation suggests that these activities can already be discriminated by static joint locations alone \citep{kuehne11iccv}. UCF50 has been extended to UCF 101 \citep{soomro12arxiv}, significantly increasing the number of categories to 101 and including 2.4 million frames at a rather low resolution of 320x240.
The Sports-1M dataset exceeds all datasets with respect to number of clips (1.1 million) and categories (487 different sports), which are, however, only weakly labeled.
Hollywood2 \citep{marszalek09cvpr}, HMDB51 \citep{kuehne11iccv}, and ASLAN \citep{kliper12pami} have very diverse activities. Especially HMDB51 \citep{kuehne11iccv} is an effort to provide a large scale database of 51 activities while reducing the database bias. Although it includes similar, fine-grained activities, such as \emph{shoot bow} and \emph{shoot gun} or \emph{smile} and \emph{laugh}, most classes have a large inter-class variability and the videos are low-resolution. ASLAN \citep{kliper12pami} focuses on a larger number of activities but with little training data per category. The task is to identify similar videos rather than categorising them.
A significantly larger video collection is evaluated during the TRECVID challenge \citep{over12tv}. The 2012 challenge consisted of 291h of short videos from the Internet Archive
(archive.org) and more than 4,000h of multi-media (audio and video) data. The challenge covers different tasks including semantic indexing and multi-media event recognition of 20 different event categories such as \emph{making a sandwich} and \emph{renovating a home}. Large parts of the data are, however, only available to the participants during the challenge.
Although our dataset is easier in respect to camera motion and background, it is challenging with respect to a smaller inter-class variability.

The datasets Coffee and Cigarettes \citep{laptev07iccv} and High Five \citep{patron10bmvc}  are different to the other movie datasets by promoting activity detection rather than classification. This is clearly a more challenging problem as one not only has to classify a pre-segmented video but also to detect (or localize) an activity in a continuous video. %
As these datasets have a maximum of four classes, our dataset goes beyond these by distinguishing a large number of classes.
The recent MPII Movie Description dataset \citep{rohrbach15cvpr} does not label clips with labels but with natural sentences which are sourced from movie scripts and audio descriptions for the blind.

The third category of datasets is targeted towards surveillance. The PETS~\citep{ferryman07pets} or SDHA2010\footnote{http://cvrc.ece.utexas.edu/SDHA2010/} workshop datasets contain real world situations from surveillance cameras in shops, subway stations, or airports. They are challenging as they contain multiple people with high partial occlusion. 
The UT interaction~\citep{ryoo09iccv} requires to distinguish 6 different two-people interaction activities, such as \emph{punch} or \emph{shake hands}. %
The VIRAT~\citep{oh11cvpr} dataset is a recent attempt to provide a large scale dataset with 23 activities on nearly 30 hours of video. Although the video is high-resolution people are only of 20 to 180 pixel height. 
Overall the surveillance activities are very different to ours which are challenging with respect to fine-grained hand motion.

Next we discuss the domain of \emph{Assisted daily living (ADL) datasets}, which also includes our dataset. The University of Rochester Activities of Daily Living Dataset (URADL)~\citep{messing09iccv} provides high-resolution videos of 10 different activities such as \emph{answer phone}, \emph{chop banana}, or \emph{peel banana}. Although some activities are very similar, the videos are produced with a clear script and contain only one activity each.
In the TUM Kitchen dataset~\citep{tenorth09iccw} all subjects perform the same composite activity (\emph{setting a table}) and rather similar actions with limited variation. 
\citet{roggen10icnss} and \citet{torre09tr} present recent attempts to provide
several hours of multi-modal sensor data (\eg body worn acceleration and object location).
But unfortunately people and objects are (visually) instrumented, making the videos visually unrealistic. %
In the CMU-MMAC dataset \citep{torre09tr} all subjects prepare the identical
five dishes with very similar ingredients and tools. In contrast to this our dataset contains \DBnTasks diverse dishes, where each subject uses different ingredients and tools in each dish. The authors also record an egocentric view. Similarly to \citep{farhadi10cvpr,fathi11iccv,stein13acm} the camera view mainly shows hands and manipulated cooking ingredients. Also recorded in an egocentric view, \citet{pirsiavash12cvpr} propose a dataset of 18 diverse daily living activities, not restricted to the cooking domain, recorded in different houses in non-scripted fashion. 

Overall our dataset fills the gap of a large database with on the one hand a detection challenge of fine-grained activities and on the other hand a recognition challenge of highly variable composite activities.

\subsection{Advances in activity recognition}
\label{sec:related:activityRecognition}

Activity recognition for still images has been advanced \eg by jointly modeling people and objects \citep{yao12tpami} or scenes and objects \citep{li07iccv}. In the following we focus on recognizing activities in video, distinguishing three aspects: holistic features for activity recognition, exploiting body pose, and modelling the temporal structure of activities.

To create a discriminative feature representation of a video, many approaches first detect space-time interest points \citep{chakraborty11iccv,laptev05ijcv} or sample them densely \citep{wang09bmvc} and then extract diverse descriptors in the image-time volume, such as histograms of oriented gradients (HOG) and histograms of oriented flow (HOF) \citep{laptev08cvpr} or local trinary patterns \citep{yeffet09iccv}.
\citet{messing09iccv} found improved performance by tracking Harris3D interest points \citep{laptev05ijcv}. The state-of-the-art Dense Trajectories approach from \citet{wang13ijcv} uses this idea: it tracks dense feature points and extracts strong video features around these tracks, namely HOG, HOF, and Motion Boundary Histograms \citep[MBH,][]{dalal06eccv}. They report state-of-the art results on several datasets including KTH \citep{schuldt04icpr}, UCF YouTube \citep{liu09cvpr}, Hollywood2 \citep{marszalek09cvpr}, and HMDB51 \citep{kuehne11iccv}.
Recently, \citet{wang13iccv} improved their approach by removing background flow and by ensuring that detected humans do not contribute to the background motion estimation. Additionally they replace the BoW encoding with Fisher vectors. The computational effort of this approach can be significantly reduced by replacing dense flow with motion information from video compression \cite{kantorov14cvpr}. 
As alternative to manually defined activity features, \citet{taylor10eccv}, \citet{baccouche11hbu}, \citet{le11cvpr}, and \citet{ji13tpami} use deep learning with convolutional neural networks to learn an activity feature representation. So far these approaches cannot reach the manually defined Dense Trajectories 
even when learning on a database of over a 1 million videos \citep{karpathy14cvpr}.

Human body poses and their motion frequently characterize human activities and interactions. %
This has been exploited in Microsoft's Kinect, which uses human pose as a game controller but relies on a depth sensor to recognize human pose \citep{shotton11cvpr}. 
Earlier work in human pose based activity recognition employed motion capture systems using physical on-body markers to reliably capture human poses, e.g. \citep{campbell95iccv}.
Such an approach is impractical for recording realistic data.
Recently a number of hand and pose-centric approaches have been proposed for activity recognition for more realistic video recordings 
\citep{fathi11iccv,packer12cvpr,yao11bmvc,sung11corr,raptis13cvpr,jhuang13iccv} as well as in
static images
\citep{yang11cvpr,yao12tpami}. \citeauthor{packer12cvpr} demonstrate
impressive results in recognition of kitchen activities using body poses
recovered from depth images. \citet{fathi11iccv} propose a hand-centric approach
for learning effective models of activities from egocentric video by observing
regularities in hand-object interactions. Hand poses have been shown to
facilitate extraction of appearance features for activity
recognition in static images \citep{karlinsky10nips}. 
Pose-based models are effective for activity recognition when body
poses can be estimated reliably, as \eg in depth
images \citep{packer12cvpr,sung11corr}.
\citet{mittal11bmvc} and \citet{gkioxari13cvpr} aim for
specialized representations for hands, but do not apply them to pose estimation
or activity recognition.
\citet{jhuang13iccv} study the benefits of pose estimation for activity recognition on a subset of the HMDB dataset \citep{kuehne11iccv}. They show that ground truth pose, estimated over time can significantly outperform the holistic Dense Trajectories features \citep{wang13ijcv}; this is also true for estimated pose using \citep{yang12pami} but only on a subset where the full body is visible.

Although several interesting techniques have been proposed to model the temporal structure of videos, they typically perform only below or on par with bag-of-word based approaches: A simple temporal structure is encoded in the template-based Action MACH from \citet{rodriguez08cvpr}, \citet{brendel11iccv} model temporal and spatial structure by segmenting the space-temporal volume, and \citet{niebles10eccv} model activities as a temporal composition of primitive actions and discriminatively learn such models. While \citeauthor{niebles10eccv} fix anchor points and the length of the temporal segments before training, \citet{tang12cvpr} learn all parameters from data using a variable-duration hidden Markov model. An AND/OR graph structure can be used to combine different features at its nodes \citep{tang13iccv} or model co-occurring and consecutive actions \citep{gupta09cvpr}. Recently \citet{pirsiavash14cvpr} have shown how to efficiently parse activity videos with segmental grammars.

\subsection{Natural language text for activity recognition}
\label{sec:related:textForActivity}

Natural language descriptions have shown beneficial for image segmentation \citep{socher10cvpr} or recognizing object categories \citep{wangME09bmvc,elhoseiny13iccv}. %
Similar to our work, \citeauthor{elhoseiny13iccv} use classifiers trained on the known classes.  Representing the text descriptions with \tfidf{} (term frequency times inverse document frequency)  vectors for relevant encyclopedic entries, they compare a regression, a domain adaptation, and a newly proposed constrained optimization formulation to learn a function from the textual vector to the visual classifier space. On two fine-grained visual recognition datasets, CU200 Birds \citep{welinder10tr} and Oxford Flower-102 \citep{nilsback08icvgip}, they show the benefit of their constraint optimization approach.
Semantic similarity from linguistic resources has also been used to allow zero-shot recognition in images via attributes and direct similarity \citep{rohrbach10cvpr} and by learning an embedding into a linguistic word vector space \citep{socher13nips,frome13nips}. Additionally to transferring knowledge one can exploit the unlabeled instances to improve recognition, assuming a transductive setting. For this, \citet{fu13pami} %
exploit the test-data distribution by performing a single round of self-training by averaging over the k-nearest neighbors. %

\citet{teo12icra} improve activity recognition by adding object detectors, which are selected based on the linguistic co-occurrence statistics in the newswire Gigaword Corpus. A similar idea is pursued by \citet{motwani12ecai}, who mine and cluster verbs from descriptions of the video snippets in the MSVD dataset \citep{chen11acl}. 
\citet{zhang11iccvw} show that \tfidf{} can identify the most relevant terms in text descriptions collected for seven video scenes allowing to yields close to perfect (98\%) recognition accuracy on their dataset.
\citet{ramanathan13iccv} jointly recognize actions and roles in YouTube videos using their captions. They mine a large number of YouTube descriptions and use a topic model to estimate the semantic relatedness between an action/role and a description.

Another line of work focuses on describing videos with natural language descriptions. Recently \citet{guadarrama13iccv} generated simple sentences for the Microsoft Video Description corpus \citep{chen11acl} containing challenging web videos. \citet{das13cvpr} compose descriptions for kitchen videos of their YouCook dataset showing YouTube cooking videos. Finally, we have shown how to learn a translation model for generating natural sentences on our dataset \citep{rohrbach13iccv}.

\subsection{Relations to our work}
\label{sec:related:dscripts:relations}

Most of the activity recognition approaches and datasets have been evaluated on full-body motion or challenging web or movie datasets but not on fine-grained motions with low inter-class variability. We therefore evaluate the holistic Dense Trajectories approach from \citet{wang13ijcv} as well as two pose-based and two hand centric approaches on our \MpiNew dataset. Our pose-based approach encodes trajectories of body joints using features motivated from the sensor-based activity recognition community \citep{zinnen09iswc}. The features are also similar to the relational and distance features defined on joints by \citeauthor{jhuang13iccv}. Similarly to their work we define relational and distance metrics between joints per frame and over time. However, our activities contain very subtle motions and the people have a very similar pose for most activities, which reduces the benefits of this feature representation.
\citeauthor{jhuang13iccv} examine the advantages of focusing Dense Trajectories \citep{wang13ijcv} on body joints. In our static scene (holistic) Dense Trajectories are already restricted to human body as the features are only extracted on moving points. However, in this work we propose to focus on hands, as they are the main cue for recognizing our fine-grained activities and participating objects.

In \citep{amin13bmvc} we improve the hand localization by leveraging multiple cameras to handle self-occlusion. In this work we remain monocular and propose to use a specialized hand detector to improve pose estimation and activity recognition.

To improve fine-grained activities and their participating objects we train a classifier on stacked classifier scores from co-occurring activities/objects as well as from temporal context after max pooling. Classifier stacking has previously been explored \eg in \citep{ting97stacked,liu12icpr,sill09arxiv}. Most relevant to our work, \citet{liu12icpr} try to optimize the usage of training data and avoid over-fitting when learning stacked video classifiers. This could be beneficial when applied to our approach.

In this work we exploit cooking instructions (script data) to extract which activities, tools, and ingredients are relevant for a certain dish (composite activity). For this we compare co-occurrence statistics with \tfidf{}, which has also been used by \citet{zhang11iccvw} and \citet{elhoseiny13iccv} to extract relevant concepts for video scene and object recognition. We find that \tfidf{} better discriminates different dishes and improves performance in most cases. 
Script data allows for zero-shot recognition, which has mainly been used for object recognition, but also for multi-media data by \citet{fu13pami}. \citeauthor{fu13pami} learn a latent attribute representation on the known classes, but then use manually defined attribute associations to transfer.

While the temporal structure, \ie temporal ordering, seems an important component to recognize activities, so far mainly the short term structure of short video clips has been explored \citep[\eg][]{gupta09cvpr,brendel11iccv,tang12cvpr}. In this work we exploit temporal co-occurrence within the same time interval and context of short actions and their participating objects within the entire video using max pooling. For long term composite activities we aggregate its components  with max pooling ignoring the temporal order. Nevertheless, we believe that the temporal structure of scripts \citep{regneri10acl} might form a good prior for the temporal structure of videos and vise-versa. \citet{bojanowski14eccv} have recently shown the benefit of movie scripts as a weak supervision. They use the ordering constraints provided by the script data to localize the actions and to learn action models.

Finally we shortly summarize how this work extends our original publications \citep{rohrbach12cvpr} and \citep{rohrbach12eccv}.  
First, we updated the dataset by correcting and unifying some of the annotations and adding a few more videos. We refer to this new version as MPII Cooking 2. It supersedes both previous datasets, see  \tableref{tbl:dataset}. Second, we present hand-centric approaches for fine-grained recognition, namely an integration of pose-estimation and hand detector and Hand centric features for activity recognition \citep[arXiv:][]{senina14arxiv}. Third, we integrated our Propagated Semantic Transfer (PST) from \citet{rohrbach13iccv} for composite recognition. Fourth, we extended qualitative and quantitative results. Fifth, we extended the discussion of related work. Sixth, we rerun experiments with updated version of Dense Trajectories \citep{wang13iccv}. And last, we will release the updated version of the dataset, new intermediate features as well as the script data.

\section{Dataset ``\MpiNew''}
\label{sec:dataset}
\begin{figure}[t]
\begin{center}
  \includegraphics[width=\columnwidth]{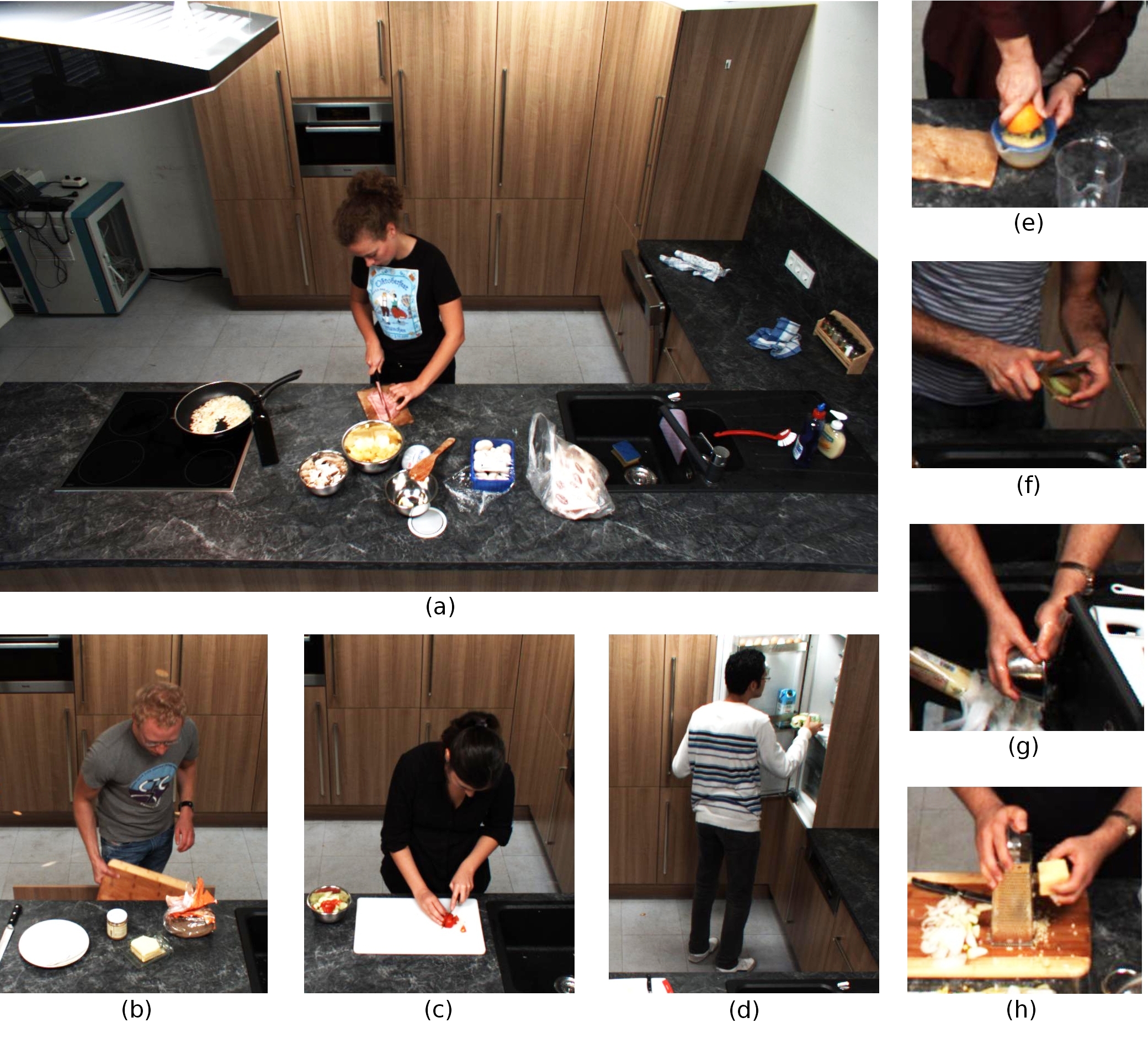}
  \caption{Single frames from the dataset depicting fine-grained cooking activities and diverse sets of tools and ingredients (participants).  (a) Full scene of \emph{slicing} in the composite activity \emph{omelet}, and crops of (b) \emph{take out}, (c) \emph{dicing}, (d) \emph{take out}, (e) \emph{squeeze}, (f) \emph{peel}, (g) \emph{wash}, (h) \emph{grate}}
  \label{fig:dataset}
\end{center}
\end{figure}

For our dataset we video-recorded human subjects cooking a diverse set of dishes, \eg \emph{making pizza} or \emph{preparing cucumber}. The dishes form the \emph{composite activities} and the individual steps taken are the \emph{fine-grained activities}, \eg \emph{cut}, \emph{pour}, or \emph{spice}. All videos have a composite label and are annotated with time intervals. Each time interval has a fine-grained activity and the participating objects as labels. A subset of frames was annotated with human pose and hands. In the following we provide details and statistics of the dataset, \Figsref{fig:teaser} and \ref{fig:dataset} show example frames of the dataset.

\subsection{Dataset statistics and versions}
\label{sec:database:stats}

We recorded \DBnSubjects subjects in \DBnVideoSeq videos with a total length of more than \DBhours hours or \DBnFrames frames. Each video contains a single subject preparing a certain dish.

The dataset was recorded in two batches. The first part contains few, but very diverse and complex dishes (see upper part of \tableref{tbl:list:dishes}) and was presented in \citep{rohrbach12cvpr}. The second part, presented in \citep{rohrbach12eccv}, focuses on composite activities and thus contains significantly more dishes/composites which are slightly shorter and simpler, see lower part of \tableref{tbl:list:dishes}. The second set of composite activities are selected according to our script corpus which we describe below in \secref{sec:mining-semrel}.
We ignored some of them which were either too elementary to form a composite activity (\eg \emph{how to secure a chopping board}), were duplicates with slightly different titles, or because of limited availability of the ingredients (\eg \emph{butternut squash}). 

For this work we corrected and unified some of the annotations and added a few more videos. We refer to this new dataset version as \MpiNew. It supersedes both previous datasets. Table \ref{tbl:dataset} compares the different versions and shows different statistics about them.
The table also shows the proposed training/vali\-dation/test split, which is selected in a way that for all 31 composite activities in the test set, there are at least 3 training/validation videos and there is no overlap between training, validation, and test subjects.
In contrast to the earlier versions we avoid multiple test splits for simpler evaluation and to reduce the computational burden for other researchers evaluating on the dataset. 

\begin{table}
\begin{tabular}{p{1cm} p{6.8cm}}
\MpiDs & sandwich, salad, fried potatoes, potato pancake, omelet, soup, pizza, casserole, mashed potato, snack plate, cake, fruit salad, cold drink, and hot drink \\
\EccvDs & \textbf{cooking pasta}, juicing \{\textbf{lime}, \textbf{orange}\}, making \{\textbf{coffee}, \textbf{hot dog}, \textbf{tea}\}, pouring beer, preparing \{asparagus, \textbf{avocado}, \textbf{broad beans}, broccoli and cauliflower, \textbf{broccoli}, carrots and potatoes, \textbf{carrots}, \textbf{cauliflower}, \textbf{chilli}, \textbf{cucumber}, \textbf{figs}, \textbf{garlic}, \textbf{ginger}, \textbf{herbs}, \textbf{kiwi}, \textbf{leeks}, \textbf{mango}, \textbf{onion}, \textbf{orange}, peach, peas, \textbf{pepper}, \textbf{pineapple}, \textbf{plum}, \textbf{pomegranate}, \textbf{potatoes}, \textbf{scrambled eggs}, spinach, spinach and leeks\}, \textbf{separating egg}, sharpening knives, \textbf{slicing loaf of bread}, using \{microplane grater, pestle and mortar, speed peeler, \textbf{toaster}, tongs\}, zesting lemon\\
\end{tabular}
\caption{Composite activities (dishes) of \MpiNew dataset, composites marked in bold are part of the test split.}
\label{tbl:list:dishes}
\end{table}

\begin{table*}[t]
 \setlength{\tabcolsep}{4pt}
\footnotesize
\begin{tabular}{l   c c c c r r r }
 \toprule %
 &   videos &  subjects & \multicolumn{2}{c}{categories}  &  \multicolumn{1}{c}{ground truth}   &  \multicolumn{1}{c}{attribute} & video  \\
 &    &                & composites & attributes& time intervals &instances  &      duration  \\
\cmidrule(r){1-1} \cmidrule(lr){2-2} \cmidrule(lr){3-3} \cmidrule(lr){4-4} \cmidrule(lr){5-5} \cmidrule(lr){6-6} \cmidrule(lr){7-7} \cmidrule(l){8-8} 
\MpiDs  \citep{rohrbach12cvpr} & 44  & 12 & 14 & 218 &  3,824 & 15,382& \  3-41 min\\
\EccvDs \citep{rohrbach12eccv} & 212 & 22 & 41 & 218 & 8,818 & 33,876 &\ 1-23 min\\
combined  & 256 & 30 & 55 & 218 & 12,642 & 49,258 &\ 1-41 min\\
\cmidrule(r){1-1} \cmidrule(lr){2-2} \cmidrule(lr){3-3} \cmidrule(lr){4-4} \cmidrule(lr){5-5} \cmidrule(lr){6-6} \cmidrule(lr){7-7} \cmidrule(l){8-8} 
\MpiNew           & \DBnVideoSeq & \DBnSubjects & \DBnTasks &   \DBnAttributes  & \DBnGtInterval & \DBnAttributeInstances & \ 1-41 min \\
 - Training set   & 201 & 24 & 58 & \DBnAttributes & 10,931& 42,619& 1-41 min\\
 - Validation set & 17  & 1  & 17 &    107 & 445 &  1,662 & 1-8 min\\
 - Test set       & 42  & 5  & 31 &  169   & 2,102 & 8,023 &1-13 min\\
\bottomrule \end{tabular}
\caption{Dataset statistics. Note that the train/val/test split do not add up to the full dataset, as some videos of the test subjects are not used as they have less than three train/val videos.}
\label{tbl:dataset}
\end{table*}

\begin{table*}[t]
\center
\begin{small}
\begin{tabular}{r@{\ }p{40mm} r@{\ }p{50mm} r@{\ }p{48mm} }
\cmidrule(r){1-2}  \cmidrule(lr){3-4} \cmidrule(l){5-6}
1. & get a large sharp knife &1.&gather your cutting board and knife. & 1.&wash the cucumber\\
2. &  get a cutting board & 2. & wash the cucumber. & 2. & peel the cucumber\\
3. & put the cucumber \newline on the board & 3. & place the cucumber flat\newline on the cutting board. & 3. & place cucumber on\newline a cutting board.\\
4. & hold the cucumber \newline in your weak hand &4. & slice the cucumber  \newline horizontally  into round  slices. &4. & take a knife and rock it \newline back and forth on the cucumber\\
5. & chop it into slices with \newline your strong hand && & 5. &make a clean thin slice each time.\\\cmidrule(r){1-2}  \cmidrule(lr){3-4} \cmidrule(l){5-6}
\end{tabular}
\end{small}
\caption{Three example scripts for the composite activity \emph{preparing cucumber}.} \label{tbl:cucumber-ex}
\end{table*}

\subsection{Datasetß recording and annotation protocol}

To record realistic behavior we neither asked subjects to perform certain activities nor to follow a certain recipe but we told them only which dish they should prepare. This resulted in a larger variety of how subjects prepared things. This means subjects used different tools for preparation (\emph{knife} or \emph{peeler} for \emph{peeling}), took different steps (\eg some people cooked the vegetables some did not), and did things in different temporal orders for the same dish (\eg \emph{washed} the vegetable before or after they \emph{peeled} it).
Before the recording the subjects were shown our kitchen and places of tools and ingredients to feel at home. During the recording subjects could ask questions in case of problems and some listened to music. We always started the recording with an empty and clean kitchen, prior to the subject entering the kitchen and ended it once the subject declared to be finished, \ie we did not include the final cleaning process. 
Most subjects were university students from different disciplines recruited by e-mail and publicly posted flyers. Subjects were paid per hour and cooking experience ranged from beginner cookers to amateur chefs.  %

Composite activities are annotated on the level of each video.
Fine-grained activities were annotated with a two-stage revision phase with start and end frame using the annotation tool Advene \citep{aubert07mm}. In addition to the activity category each annotation consists of used tools, ingredients, and locations (we refer to them as participants).
Composite activities were chosen as described in \secsref{sec:database:stats} and \ref{sec:mining-semrel}. Activity, tool, ingredient, and location categories were chosen to describe all activities the human subjects were performing. The decision was made after the recording on the base what the human subjects did. With respect to the level of detail, we do not annotate the specific motions (\eg move arm up or down) but what effect or semantic they have (\eg open versus close). See \tableref{tbl:results:fine-grained-activities} for the chosen granularity.

We recorded  in our kitchen (see \figref{fig:dataset}(a)) with a 4D View Solutions system using a Point Grey Grasshopper camera with 1624x1224 pixel resolution at 29.4fps and global shutter. The camera is attached to the ceiling, recording a person working at the counter from the front. 
We  provide the sequences as single frames (jpg with compression set to 75) and as video streams (compressed weakly with mpeg4v2 at a bit-rate of 2500).
For most videos we recorded 7 additional camera views on the kitchen, a subset was used and released by \citet{amin13bmvc}. Although they are not used in this work we will make the remaining 7 views available upon publication. All fine-grained and composite activity annotations are also valid for the other cameras as each frame was synchronized across all 8 cameras.

We also provide intermediate representations of holistic video descriptors, human pose detections, tracks, and features defined on the body pose. We hope this will foster research at different levels of activity recognition.

The dataset provides furthermore human body pose annotations (see \secref{sec:poseChallenge}), script data (see \secref{sec:mining-semrel}) and there exist textual descriptions in the TACoS \citep{regneri13tacl} and TACoS multi-level corpus \citep{rohrbach14gcpr}. The descriptions in TACoS describe what happens in a specific video and are temporally aligned to the video, \ie they provide a textual annotation. In contrast, the scripts used in this work are collected independently of the video and thus contain domain or script knowledge, \ie what activities and what objects are likely used for a certain dish. As they are not specific to the training videos they allow to transfer and generalize to novel test scenarios.

\subsection{Pose Challenge} 
\label{sec:poseChallenge}
A subset of frames have articulated human pose and hand annotations to learn and evaluate pose estimation approaches and hand detectors.
For human pose we annotated the frames with right and left shoulder, elbow, wrist, and hand joints as well as head and torso.
We have 2,994 frames of 10 subjects for training  
of pose annotation and an additional of 4,250 training images with hand points used for training the hand detector. For testing we sample 1,277 frames from all activities with 7 subjects as test set for the pose challenge.
All training and test frames are from MPII Cooking \citep{rohrbach12cvpr} and thus avoid an overlap with the test subjects and test composites in \MpiNew.

\subsection{Mining script data for composite activities}
\label{sec:mining-semrel}

Linguistics and psychology literature knows prototypical sequences of certain activities as so-called \emph{scripts} \citep{schank77book,barr82book}. Scripts describe a certain scenario which corresponds to composite activities in our case. Scenarios (\eg \emph{eating in a restaurant}) are temporally ordered events (\emph{the patron enters restaurant, he takes a seat, he reads the menu,...}) and subjects (\emph{patron, waiter, food, menu,...}).  Written event sequences for a scenario can be collected on a large scale using crowd-sourcing \citep{regneri10acl}. We make use of this method to collect scripts for our composite activities and assembling a large number of written sequences for each of those.

We collect natural language sequences similar to \citet{regneri10acl} using Amazon's Mechanical Turk\footnote{http://www.mturk.com}. For each composite activity, we asked the subjects to give tutorial-like sequential instructions for executing the respective kitchen task. The instructions had to be divided into sequential steps with at most 15 steps per sequence. 
We select 53 relevant kitchen tasks as composite activities by mining the tutorials for basic kitchen tasks on the webpage 
``Jamie's Home Cooking Skills''\footnote{{http://www.jamieshomecookingskills.com}}.  
All those tasks/scenarios are about processesing ingredients or using certain kitchen tools.
In addition to the data we collected in this experiment, we use data from the OMICS corpus \citep{omcs2} and \citet{regneri10acl} for 6 kitchen-related composite activities.  This results in a corpus with 59 composite activities and 2,124 sequences in sum, having a total of 12,958 individual event descriptions. Note that for practical reasons we only recorded videos for 35 of these composite activities as discussed in \secref{sec:database:stats}. They are listed in \tableref{tbl:list:dishes} under ``MPII Composites''.

This script corpus provides much more variation than the limited number of video training examples can capture. Of course this also poses a challenge, because we need to overcome the problem of different wordings and coordinated events:
\tableref{tbl:cucumber-ex} shows three examples we collected for the composite activity \emph{preparing cucumber}. They differ in verbalization (\eg \emph{slice}, \emph{chop}, and \emph{make a slice}) and granularity (\emph{getting} something is often left out). Further, the sequences reflect different ways of preparing the vegetable, some include \emph{peeling} it, some do not \emph{wash} it, and so on. Some sentences contain conjugated events (\emph{take a knife and rock it...}). While we clean the data to a certain degree by fixing spelling mistakes and resolving pronouns with the method from \citet{bloem-regneri-thater:KONVENS}, we end up with both challenges and blessings of a noisy but big script corpus.
 
In \secref{sec:semrel} we will describe how we extract semantic relatedness from this data.

\section{Hand detection and pose estimation}
\label{sec:approach:handAndpose}
One goal of this paper is to investigate the applicability of state-of-the-art pose estimation methods in the context of
activity recognition. Therefore, in this section we propose our new pose estimation method based on \citet{andriluka11ijcv} and benchmark it on our dataset together with state-of-the-art pose estimation methods.
Another goal is to demonstrate the importance of hand-based features for recognizing activities and their participants. For this we need to localize hands, which is in itself a challenging task due to partial occlusions, obstruction by manipulated objects, and variability of hand postures.
In order to achieve high quality hand localization we leverage two complementary sources of information. We exploit the characteristic appearance of hands in order to train an effective hand detector.
We then integrate observations from this detector in our pose estimation approach to take advantage of the context provided by the other body parts.
As another finding, we show that localization of all body parts benefits significantly from our specialized hand detector. 

In the following we introduce our hand detector (\secref{sec:approach_pose:hand_detection}) and pose estimation method (\secref{sec:approach:pose}) as well as how we combine them (\secref{sec:approach_pose:handinpose}). In \secref{sec:pose_results} we evaluate our proposed approaches as well as state-of-the-art pose estimation methods on our dataset.

\subsection{Hand detection based on local appearance}
\label{sec:approach_pose:hand_detection}
As a basis for our hand detector we rely on the deformable part models \citep[DPM,][]{Felzenszwalb2010PAMI}. We discuss several design choices in order to achieve best performance.

\begin{figure}[t]
\centering
{\includegraphics[scale=0.5]{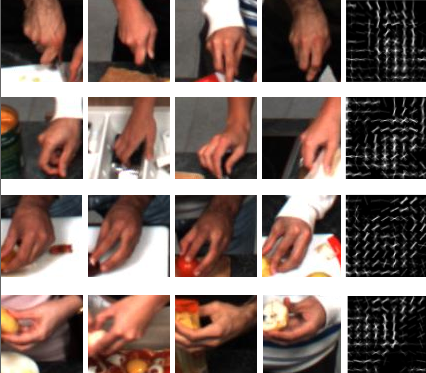}}
\caption{Examples of training images assigned to 4 different hand components,
  each row shows images from one component. Rows 1 and 2 correspond to
  right hand components, and rows 3 and 4 to left hand components.}
\label{fig:dpm_clusters}
\end{figure}

\myparagraph{Detection of left and right hands.}
We aim for a hand detector that can correctly distinguish the left and right hand of a person. The rationale behind this is that for many activities left and right hands have different roles (\eg for a cutting activity the dominant hand is typically holding a knife while the supporting hand is holding the object that is being cut). Further, we would like to avoid situations when two strong hypotheses for one of the hands are chosen over two hypotheses for both hands. We achieve this by dedicating separate DPM components to left and right hands and jointly training them within the same detector (see examples in Figure \ref{fig:dpm_clusters}). Note that in contrast to the default setting mirroring is switched off in DPM. 
At test time we pick the best scoring hypothesis among the components corresponding to left and right hands.

\myparagraph{Component initialization.}
We capture the variance of hand postures by decomposing the hands' appearance
into multiple modes and representing each mode with a specific DPM component.
We found that a rather large number of components is necessary to achieve good
detection performance. 
We initialize the components by clustering the HOG descriptors of the training
examples using K-means as in \cite{Divvala:2012:HIA}. The detection further
improves by first clustering the training examples by hand orientation and then
by HOG. 

\myparagraph{Body context.}
We improve the hand localization by augmenting the hand detector with the
context provided by a person detector. We rely on the person detector to constrain the search for
hands to the image locations within the extended person bounding box and also
constrain the scale of the hands detector to the scale of the person hypothesis.

\subsection{Pose estimation}
\label{sec:approach:pose}

\newlength{\trimoffsetx}
\setlength{\trimoffsetx}{2cm}

\newlength{\trimoffsety}
\setlength{\trimoffsety}{0cm}

\begin{figure*}[t]
\centering
\subfigure[]{
\newcommand{\midruleBM}{\cmidrule(lr){1-1}  \cmidrule(lr){2-7} \cmidrule(lr){8-8}}
{\begin{minipage}{0.5\linewidth}
\begin{scriptsize}
\begin{center}
\setlength{\tabcolsep}{1pt}
\begin{tabular}{lrrrrrrr}
\toprule
 &   &  & \multicolumn{2}{c}{upper arm} & \multicolumn{2}{c}{lower arm}  & \\
Method &  Torso & Head &  \multicolumn{1}{c}{r} & \multicolumn{1}{c}{l} & \multicolumn{1}{c}{r} & \multicolumn{1}{c}{l}  & All\\
\midruleBM
\multicolumn{5}{l}{\textbf{Original models}}\\
CPS~\cite{Sapp10cascadedmodels}                   & {67.1}    & 0.0             & 53.4            & 48.6             & {47.3}       & 37.0               & 42.2\\
FMP~\cite{confcvprYangR11}                        & 63.9             & {72.1}   & {60.2}   & {59.6}    & 42.1                & {46.7}      & {57.4}\\
PS~\cite{Andriluka:2009}                          & 58.0             & 45.5            & 50.5            & 57.2             & 43.3                & 38.8               & 48.9\\
\midruleBM
\multicolumn{5}{l}{\textbf{Trained on our data}}\\
FMP~\cite{confcvprYangR11}                        & 79.6             &	67.7           &	60.7           & 60.8	            &	50.1              	&	50.3	             & 61.5\\
PS~\cite{Andriluka:2009}                          & 80.1    		 & 80.0 		   &67.8 & 69.6   & 48.9                & 49.6               & 66.0\\
\midruleMPI
FPS                         & 78.5  			 & 79.4  & 61.9    
& 64.1             & 62.4       & 61.0      & 67.9\\
FPS + data    & 79.3             &85.0 & 64.3                  & 64.6                     & 60.0                   & 59.8                    &    68.8\\
FPS + data + hand det                                      & 79.6                     & 
84.9                & 70.9                  & 70.0                      & 73.5                   & 70.2                    &   74.9\\
FPS + data + color                                         & 80.7                    
& 85.8  	          & 69.1                  & 67.4                     & 69.3                    & 65.5                    &    73.0\\
FPS + data + hand det + color                               & \textbf{81.3}       & \textbf{86.1}   & \textbf{72.4}    & \textbf{71.3}     & \textbf{74.4}     & \textbf{70.3}     & \textbf{75.9}\\
\bottomrule
\end{tabular}
\end{center}
\label{tab:mpi_pose_pcp}
\end{scriptsize}
\end{minipage}}
}
\subfigure[]{
{\begin{minipage}{0.45\linewidth}
{\includegraphics[width=1.0\linewidth]{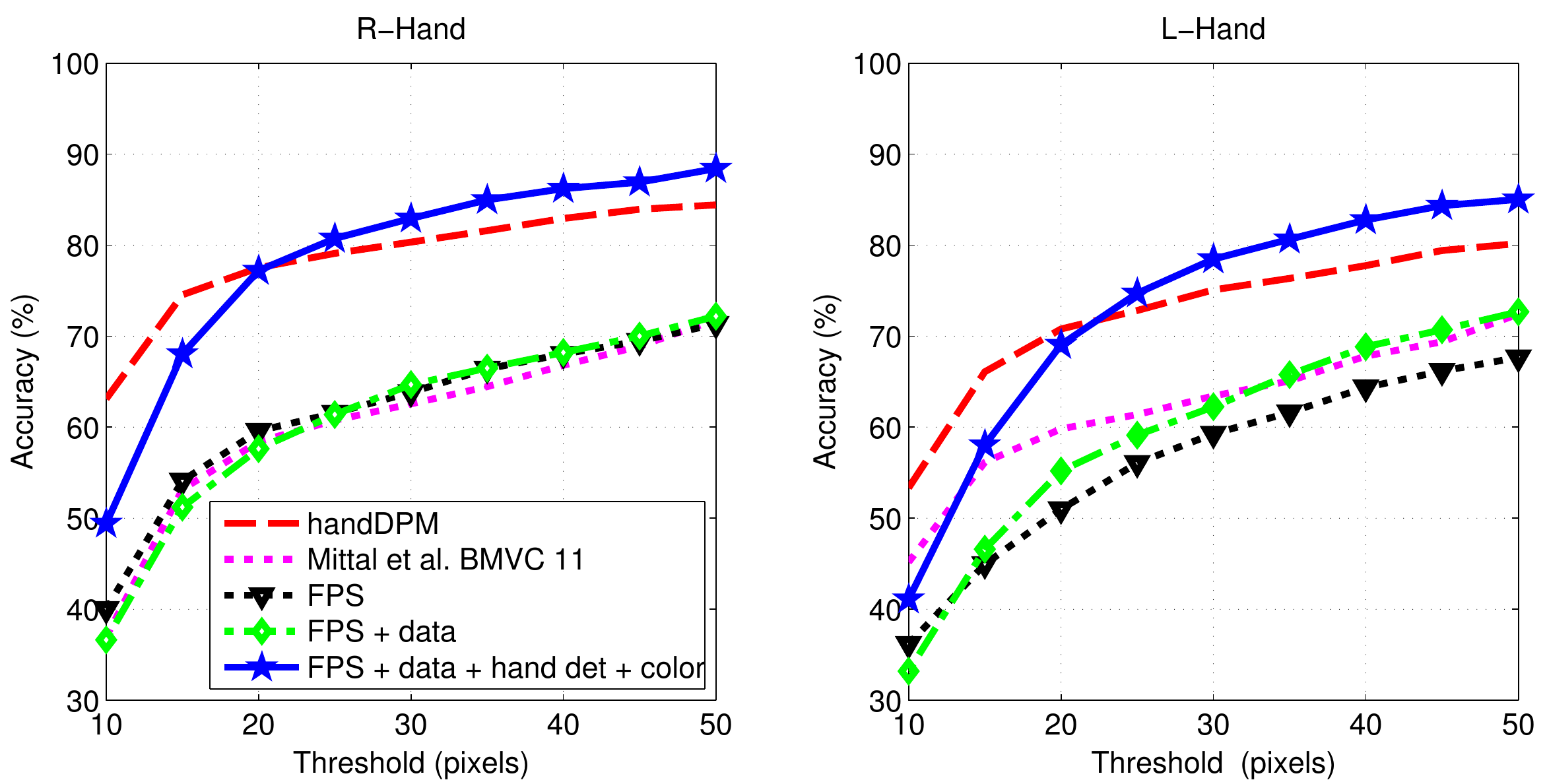}}
\label{fig:hand_pose_euc}
\end{minipage}}
}
\caption{ (a) 2D upper body pose estimation results on the ``Pose Challenge''
   of our dataset. The numbers
  correspond to the ``percentage of correct parts'' (PCP).  
  (b) Accuracy of different methods for detection of right and left hands for a varying
  distance (in pixels) from the ground truth position.
}
\end{figure*}
We base our pose estimation approach on the pictorial structures (PS) approach
\citep{Fischler1973TC,Felzenszwalb:2005:PSO}. In PS the body is represented as a collection of
rigid parts linked via a set of pairwise part relationships. Unlike the original model we define a flexible variant of the PS model (FPS) that
consists of $N=10$ parts corresponding to head, torso, as well as left and right shoulders, elbows, wrists and hands. Denoting the configuration of parts as $L =
{l_1, \ldots, l_{N}}$, and image observations as $D$, the posterior over the
part configuration is given by
\begin{equation}
\label{eq:ps}
p(L|D) \propto \prod_{(i,j) \in E} p(l_i|l_j) \cdot \prod_{i=1}^{i=N} p(D|l_i),
\end{equation}
\noindent where $E$ is a set of connected part pairs. We build on the publicly
available PS implementation from \cite{andriluka11ijcv}. 
In this model the pairwise connections between parts form a tree
structure, which permits efficient and exact inference. The pairwise terms
represent the spatial relationships between part positions and are modeled as
Gaussians with respect to relative position and orientation of parts. The
appearance of individual parts is represented with boosted part detectors and
shape context image features. 
Conceptually the formulation of \cite{andriluka11ijcv} is similar to flexible mixture of parts model \citep[FMP,][]{confcvprYangR11}. The FMP model represents appearance of each body part with a set of HOG templates. Pairwise terms are adapted depending on the particular template. Parameters of appearance templates and pairwise terms of the FMP model are jointly trained using max-margin objective. The model of \cite{andriluka11ijcv} relies on a single appearance template for all parts. Parameters of pairwise terms are estimated using maximum likelihood independently from appearance terms.
We extend this model by incorporating color features into the part likelihoods by stacking them with shape context features prior to part detector training. We encode the color as a
multidimensional histogram in RGB space using $10$ bins for each color
dimension which results in $1000$ dimensional feature vectors. We then
concatenate color and shape context features and train boosted part detectors
for each part using the combined representation. We use standard AdaBoost
for training and rely on the same weak learners as in \cite{andriluka11ijcv}.

\subsection{Combining hand detection and pose estimation}
\label{sec:approach_pose:handinpose}
We extend the image observations in Eq.~\ref{eq:ps} with detection hypotheses
for left and right hands, which we obtain using the corresponding components of
our hand detector. We denote the set of hand hypotheses produced by our hand
detector by $H = \{(d_k, s_k)|k=1,\ldots,K\}$, where $d_k$ is the image position
and $s_k$ the detection score. Based on this sparse set of detections we obtain
a dense likelihood map for the hand part $l_h$ using a kernel density
estimate:
\begin{equation}
p(H|l_h) = \sum_{k=1}^Kw_k \exp( -\sigma^2\|d_k -l_h\|^2),
\end{equation}
\noindent where $w_k = s_k - m$ is a positive weight associated with each hand
hypothesis computed by shifting the detection score by the minimal score value
$m$. There is no specific upper/lower bound for the scores $s_k$, but since DMP relies on SVM formulation the scores tend to be centered around 0 with confident negative examples having score less than -1. In practice we set $m = -1$  and ignore all detections with a smaller score than $m$.

\subsection{Evaluation: pose estimation and hand detection}
\label{sec:pose_results}
We first evaluate the results on the upper-body pose estimation task.
In order to identify the best 2D pose estimation approach we use our 2D body joint annotations (see \secref{sec:poseChallenge}). 
For evaluating these methods we adopt the PCP measure (percentage of correct parts) proposed by \citet{Ferrari:2008:PSS}. The results are shown in Figure \ref{tab:mpi_pose_pcp}.
The first three lines compare three state-of-the-art methods: the cascaded pictorial structures \citep[CPS,][]{Sapp10cascadedmodels}, the flexible mixture of parts model \citep[FMP,][]{confcvprYangR11} and the implementation of pictorial structures model \citep[PS,][]{andriluka11ijcv}, using their published pose models.
Lines 4 and 5 show the models of \citeauthor{confcvprYangR11} and \citeauthor{andriluka11ijcv} retrained on our data. Overall the model of \citeauthor{andriluka11ijcv} performs best, achieving 66.0 PCP for all body-parts. 
We attribute the improvement of PS over FMP to the following. The FMP model encodes different orientation of parts via different appearance templates, whereas the PS model uses a single template that is rotation invariant and is evaluated at all orientations. The FMP model has a larger number of parameters because appearance templates are not shared across different  part orientations. A larger number of parameters means that it is easier to overfit the FMP model than the PS model. This could explain the performance differences after retraining on our data. It could also be that finer discretization of body part orientations in the PS model compared to the FMP model is important for good performance.
As described above we base our model (FPS) on PS, adding to it flexible part configuration.

The bottom part of the Figure \ref{tab:mpi_pose_pcp} shows that this as well as our other improvements (more training data comparing to \cite{rohrbach12cvpr}, color features, and hand detections) in the model each helps to improve performance. Overall, compared to PS, we achieve an improvement from 66.0 to 75.9 PCP and most notably an improvement from 48.9 to 74.4 and from 49.6 to 70.3 for lower arms, which are most important for recognizing hand-centric activities.
We also would like to point to the benefit which hand detectors have to pose estimation (compare line 7 vs 8 and 9 vs 10).

Next we discuss the hand detection results. Our final hand detector \emph{handDPM} is based on $32$ components with $16$ components allocated to each of the hands. The
components are initialized by first grouping the training examples of each hand
into $4$ discrete orientations, and then clustering their HOG descriptors.
In the experiments on hand localization we use a metric that reflects the localization
accuracy and measures the percentage of hand hypotheses within a given distance
from the ground truth. We visualize the results by plotting the localization
accuracy for a range of distances.

\begin{figure}[t]
\centering
{\includegraphics[width=\columnwidth]{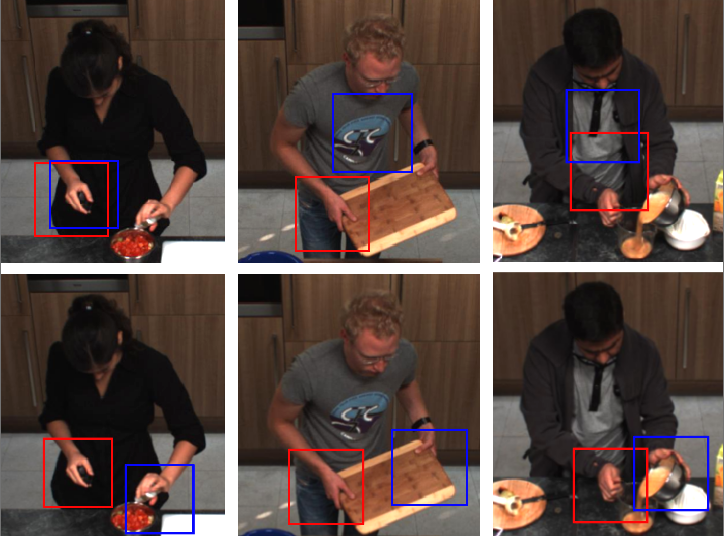}}
\caption{Pose helps to resolve failure cases of hand localization (upper row - handDPM, lower row is FPS+data+hand det+color).}
\label{fig:dpm_improvements}
\end{figure}

\figref{fig:hand_pose_euc} presents the evaluation of the localization
accuracy of both hands. We observe that our hand detector (handDPM, red-dashed curve) alone already significantly improves over the proposed FPS approach (black-dotted-triangles). 
The performance further improves when hand detection hypotheses are integrated within the pose estimation model (blue-solid-stars). However, the improvement is moderate, likely because the pose estimation
approach is not optimized specifically for hand detection and has to compromise
between localization of hands and other body parts. Some qualitative examples are shown in \figref{fig:dpm_improvements}.

We also compare our hand detector to a state-of-the-art hand detector of \cite{mittal11bmvc} using the code made publicly available
by the authors. We perform the best-case evaluation and assign the hand
hypothesis returned by the approach to the closest left
and right hand in the ground-truth, as the hand detector 
does not differentiate between left and right hands. For a fair comparison we also
filter the hand detections of \cite{mittal11bmvc} at irrelevant scales and image
locations using body context as explained before. Our
detector significantly improves over the hand detector of \cite{mittal11bmvc}, which in addition to hand appearance  also
relies on color and context features, whereas our hand detector uses hand regions
only.
Note that there are significant differences between localization accuracy of
left and right hands. We attribute this to the fact that the majority of people
in our database are right handed. Since people perform many activities with their dominant hand, the pose of the 
right hand is more likely to be constrained by various activities due to 
the use of tools such as a knife or peeler. The left hand's pose is far less deterministic and the hand is often occluded behind the counter or while holding various objects.

\section{Approaches for fine-grained activity recognition and detection}
\label{sec:approach:finegrained}
In this section we focus on fine-grained activity recognition to approach the challenges typical \eg for assisted daily living. Along with the activities we want to recognize their participating objects.
To better understand the state-of-the-art for this challenging task we benchmark three types of approaches on our new dataset. 
The first type (\secref{sec:approach:finegrained:pose}) uses features derived from upper body model motivated by the intuition that human body configurations and human body motion should provide strong cues for activity recognition. For body pose estimation we rely on our approach described in Sections \ref{sec:approach:pose} and \ref{sec:approach_pose:handinpose}.
The second type (\secref{sec:approach:finegrained:holistic}) are the state-of-the-art Dense Trajectories~\citep{wang13ijcv} which have shown promising results on various datasets. It is a holistic approach in a sense that it extracts visual features on the entire frame.
As the third type (\secref{sec:approach:finegrained:hand}) we present our hand-centric visual features, targeted at recognizing our hand-centric activities and the participating objects which are typically in the hand neighbourhood. For this we propose a hand detector (Sections \ref{sec:approach_pose:hand_detection}, \ref{sec:approach_pose:handinpose}).
Finally, we discuss our approaches to activity classification and detection in \secref{sec:approach:finegrained:det}.

\subsection{Pose-based approach}
\label{sec:approach:finegrained:pose}

Pose-based activity recognition approaches were shown to be effective using inertial sensors \citep{zinnen09iswc}. 
Inspired by \citet{zinnen09iswc} we build on
a similar feature set, computing it from the temporal sequence of 2D body configurations.

We employ a person detector \citep{Felzenszwalb2010PAMI} and estimate
the pose of the person within the detected region with 50\% border around.
This allows us to reduce the complexity of the pose estimation and simplifies the search to a single scale.
To extract the trajectories of body joints we rely on search space reduction \citep{Ferrari:2008:PSS} and tracking. 
To that end we first estimate poses over a sparse set of frames (every 10-th frame in our evaluation) and then track over a fixed
temporal neighborhood of 50 frames forward and backward. For tracking we match SIFT features for each joint separately across consecutive frames. To discard outliers we find the largest
group of features with coherent motion and update the joint position based on
the motion of this group. 
This approach combines the generic appearance model learned at training time with the specific appearance (SIFT)
features computed at test time. 

Given the body joint trajectories we compute two different feature representations. First is a manually defined statistics over the body model trajectories, which we refer to as \emph{body model features} (BM). Second is Fourier transform features (FFT) from \citet{zinnen09iswc}, which have shown effective for recognizing activities from body worn wearable sensors.

\myparagraph{Body model features (BM).}
For the BM features we compute the \emph{velocity} of all joints (similar to gradient calculation in the image domain). We bin it in an 8-bin histogram according to its direction, weighted by the speed (in pixels/frame). This is similar to the approach by \citet{messing09iccv} which additionally bins the velocity's magnitude. We repeat this by computing \emph{acceleration} of each joint. Additionally we compute \emph{distances} between the right and corresponding left joints as well as between all 4 joints on each body half. Similar to the joint trajectories (\ie trajectories of x,y values) we build corresponding ``trajectories'' of distance values by stacking the values over temporally adjacent frames.
For each distance trajectory we compute statistics (mean, median, standard deviation, minimum, and maximum) as well as a rate of change histogram, similar to velocity. Last, we compute the angle trajectories at all inner joints (wrists, elbows, shoulders) and use the statistics (mean etc.) of the angle and angle speed trajectories. This totals to 556 dimensions.

\myparagraph{Fourier transform features (FFT).}
The FFT feature contains 4 exponential bands, 10 cepstral coefficients, and the spectral entropy and energy for each x and y coordinate trajectory of all joints, giving a total of 256 dimensions.

\myparagraph{Feature representation.}
For both features (BM and FFT) we compute a separate codebook for each distinct sub-feature (\ie\ velocity, acceleration, exponential bands etc.) which we found to be more robust than a single codebook. We set the codebook size to twice the respective feature dimension, which is created by computing k-means from all features (over 80,000).
We compute both features for trajectories of length 20, 50, and 100 (centered at the frame where pose was detected) to allow for different motion lengths. The resulting features for different trajectory lengths are combined by stacking and give a total feature dimension of 3,336 for BM and 1,536 for FFT.

\subsection{Holistic approach}
\label{sec:approach:finegrained:holistic}
Most approaches for activity recognition are based on a bag-of-words representations. We pick the state-of-the-art Dense Trajectories approach \citep{wang11cvpr,wang13ijcv} which extracts histograms of oriented gradients (HOG), flow \citep[HOF][]{laptev08cvpr}, and motion boundary histograms \citep[MBH][]{dalal06eccv} around densely sampled points, which are tracked for 15 frames by median filtering in a dense optical flow field. The x and y \emph{trajectory} speed is used as a fourth feature. Using their code and parameters which showed state-of-the-art performance on several datasets we extract these features on our data.  
Following \citet{wang13ijcv} we generate a codebook for each of the four features of 4,000 words using k-means from over a million sampled features.

\subsection{Hand-centric approach}
\label{sec:approach:finegrained:hand}

In domains where people mainly perform hand-related activities it seems intuitive to expect that hand regions contain important and relevant information for recognizing those activities and the participating objects.
Thus, in addition to using the holistic and pose-based features, we suggest to focus on the hand regions.
To obtain the hand locations we rely on our hand detector described in \secref{sec:approach_pose:hand_detection} as well as on the pose estimation method with integrated hand candidates (\secref{sec:approach_pose:handinpose}). In order to increase the robustness of the method we use both location candidates (provided by the handDPM detector and the final pose model) and sum the obtained features. 

\myparagraph{Hand-Trajectories}
We want to represent different type of information: hand motion, hand shape, and shape variations over time, as well as the appearance of objects manipulated by the hands.
We propose to densely sample the neighborhood of each hand and to track those points over time. 
For tracking and also representing the point trajectories with powerful features we adapt the approach of \cite{wang13ijcv}. 
We focus only on densely sampled points around the estimated hand positions instead of sampling the entire video frame. 
We specify a bounding box around each hand detection and densely sample points inside of it. In our experiment we use 120$\times$140 pixels bounding box around hands to include the information about the hands' context. We use 8 pixels grid spacing for points sampling and finally we get 136 interest point tracks for each frame. After extracting the features along computed tracks we create codebooks that contain 4000 words per feature.
\myparagraph{Hand-cSift}
Color information is another important cue for recognizing activities and even more prominent for recognizing the participating objects. 
Similar to the previous approach we densely sample the points in the hands' neighborhood and extract color Sift features on 4 channels (RGB+grey). We quantize them in a codebook of size 4000.

\subsection{Fine-grained activity classification and detection}
\label{sec:approach:finegrained:det}
\paragraph{Activity classification}
Given a long video we assume that it consists of multiple time intervals. Each such interval $t$ depicts a single fine-grained activity and its participating objects (e.g. \emph{dry, hands, towel}). In the following we refer to both, activities and participants, as activity attributes $a_i, (i \in \{1,\ldots,n\})$, \ie $a_i$ can be any attribute including \emph{cut}, \emph{knife}, or \emph{cucumber}. We train one-vs-all SVM classifiers on the features described in the previous sections given the ground truth intervals and labels. The classifiers provide us with real valued confidence score functions $f^{base}_i:\mathbb{R}^N\mapsto \mathbb{R}$ for attribute $a_i$ and feature vectors of dimension $N$. Combining different features is achieved by concatenating, \ie stacking, the corresponding feature vectors.
\paragraph{Activity detection}
While we use ground truth intervals for training the activity classifiers, we use a sliding window approach to find the correct interval of detection. To efficiently compute features of a sliding window we build an integral histogram over the histogram of the codebook features. We use non maximum suppression over different window lengths and start with the maximum score and remove all overlapping windows.
In the detection experiments we use a minimum window size of 30 with a step size of 6 frames; we increase window and step size by a factor of $\sqrt{2}$ until we reach a window size of 1800 frames (about 1 minute). Although this will still not cover all possible frame configurations, we found it to be a good trade-off between performance and computational costs.

\section{Modeling composite activities}
\label{sec:approach}
\begin{figure*}[t]
\center
\subfigure[Activity attribute recognition using \textbf{con}textual and \textbf{co-occ}urrence attributes vectors\label{fig:contextCooccurence}.]{\includegraphics[width=0.47\textwidth]{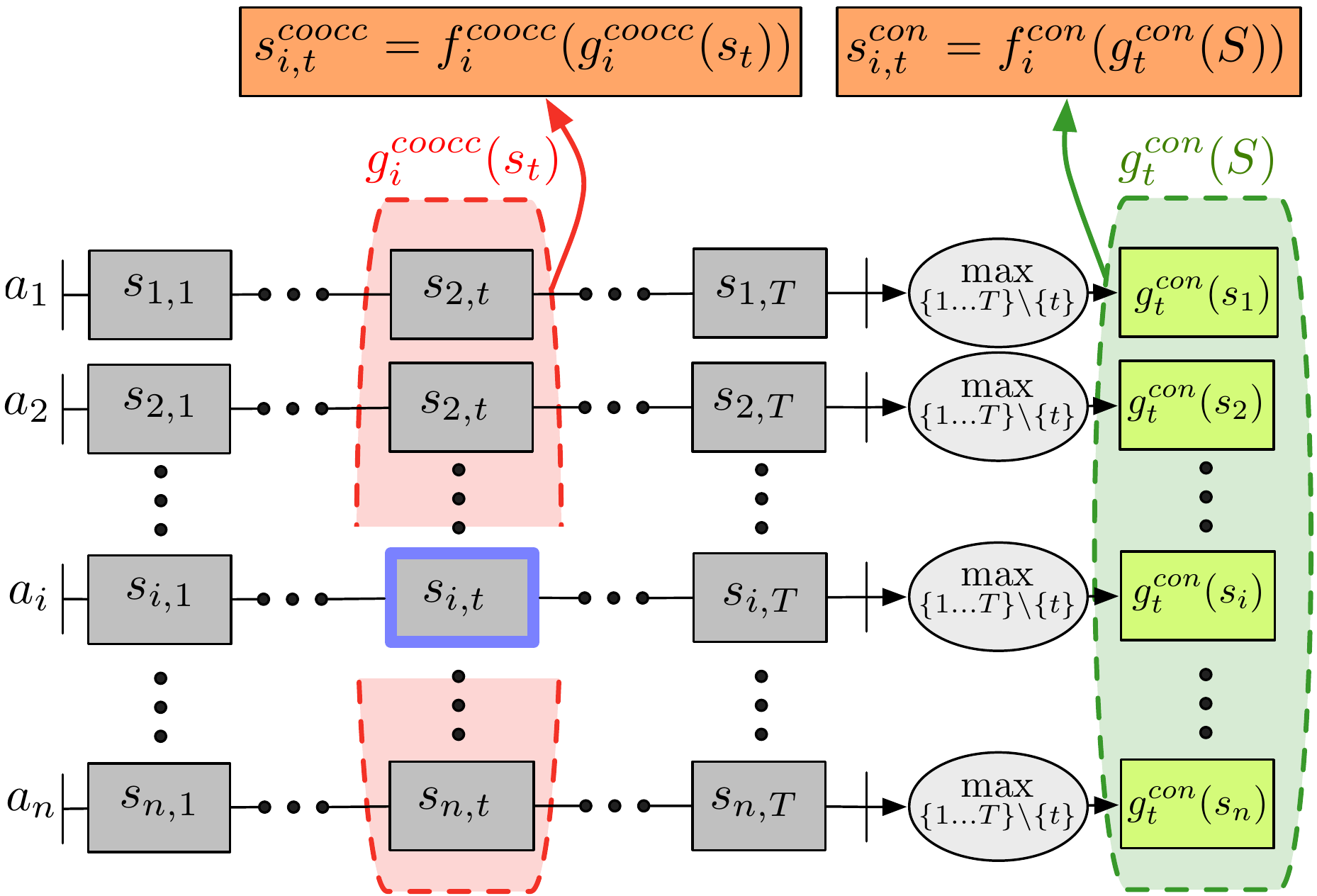}
}
\hspace{0.3cm}
\subfigure[Composite activity classification using max-pooled activity attributes\label{fig:seqClassification}.]{
\includegraphics[width=0.47\textwidth]{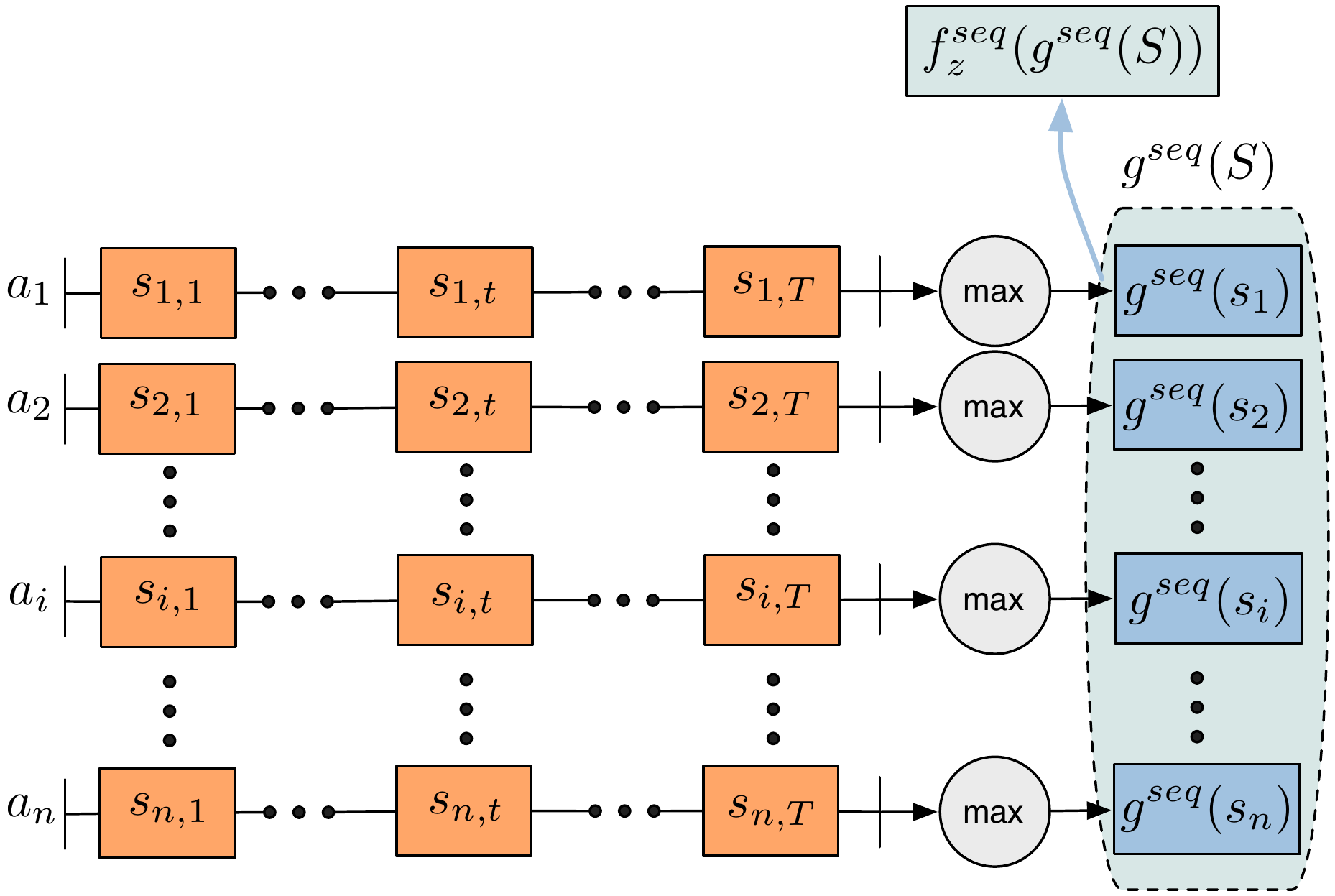}
}
\caption[Our approach to recognition of attributes and composite activities.]{Our approach to recognition of attributes (a) and composite activities (b).}
\label{fig:approachAttributesComposites}
\end{figure*}

In the previous section we discussed how we recognize fine-grained activities (such as \emph{peeling} or \emph{washing}) and their object participants (such as \emph{grater}, \emph{knife}, or \emph{cucumber}). Now we focus on exploiting the temporal context and on recognizing different composite activities, \eg \emph{preparing a cucumber} or \emph{cooking pasta}. 

For this, we first show how we exploit temporal context and co-occurrence to improve the recognition of fine-grained activities and their object participants (\secref{model:contextCooccurence}). Then, we model composite activities as a flexible combination of attributes, where attributes refer jointly to the fine-grained activities and their object participants (\secref{sec:model:task}). 
We then show how to use prior knowledge (\secref{sec:model:semrel}) to improve the recognition of composite activities, overcoming the notorious lack of training data and handling the large variability of composite activities. In \secref{sec:semrel} we discuss how to mine the semantic relatedness from script data.
Finally, in \secref{sec:model:segmentation} we introduce an automatic approach to temporal video segmentation, which removes the necessity to manually annotate the ground truth intervals in a video. 

\subsection{Recognizing activity attributes using context and co-occurrence}
\label{model:contextCooccurence}

For a time interval $t$ we want to classify if a particular fine-grained activity and its participants are present. We refer to activities and participants as activity attributes $a_i$. We distinguish three types of attribute classifiers. The first type of is 
given by the classifiers introduced in the previous section providing us with confidence score functions $f^{base}_i:\mathbb{R}^N\mapsto \mathbb{R}$ for each attribute $a_i$. Let us denote the score of a given feature vector $x_t$ at time interval $t$ as:
\begin{equation}
	s_{i,t} = f^{base}_i(x_t).
\end{equation} 
Together these score constitute a matrix $S$ of dimensions $n$  $\times$ $T$ (\# attributes $\times$ \#timestamps).
Based on these scores, we define features for context (in the same video sequence) as well as features for co-occurrence of other attributes (in the same time interval $t$). 

Contextual features formalize the intuition that adjacent time frames have strongly related attributes: \eg if a \emph{cucumber} is \emph{peeled} in one time interval, then \emph{cutting} the \emph{cucumber} is probably also present in the same video sequence. %
As visualized in \figref{fig:contextCooccurence} we define a context feature $g^{con}_t:\mathbb{R}^{n\times T}  \mapsto  \mathbb{R}^{n}$ at time $t$ by max pooling the scores of each attribute over all time intervals except $t$:  
\begin{equation}
g^{con}_t(S)=\max_{u\in\{1,...,T\}\setminus\{t\}}s_{u}
 \end{equation}
\noindent where $\max$ is an element-wise operator over all columns $s_u \in \mathbb{R}^n$ of matrix $S$.

Similarly, activity attributes happening at the same time interval $t$ are related, \eg if we \emph{peel} something it is more likely to observe also \emph{carrot} or \emph{cucumber} rather than \emph{cauliflower}. We thus define the co-occurrence as a feature $g^{coocc}_{i}:\mathbb{R}^{n}  \mapsto  \mathbb{R}^{n-1}$ by stacking all attribute scores at time $t$ excluding $s_{i,t}$:
\begin{equation}
g^{coocc}_{i}(s_t)=[s_{1,t};...;s_{i-1,t};s_{i+1,t};...;s_{n,t}],
\end{equation}
where $s_t \in \mathbb{R}^n$ is a column of matrix $S$.

Based on these features we train activity attribute SVM classifiers using the features individually or by stacking them.
Specifically we obtain corresponding confidence score functions for context: $f^{con}_i:\mathbb{R}^{n} \mapsto \mathbb{R}$ and co-occurrence: $f^{coocc}_i:\mathbb{R}^{n-1} \mapsto \mathbb{R}$, where $i$ denotes that a separate function for each attribute $a_i$ is trained.
We define corresponding scores as:
\begin{equation}
	s^{con}_{i,t} = f^{con}_i(g^{con}_t(S))
\end{equation} 
 and
\begin{equation}
	s^{coocc}_{i,t} = f^{coocc}_i(g^{coocc}_{i}(s_t)).
\end{equation}
This formulation can be easily extended to other attribute representations depending on the task and available features.

\subsection{Composite activity classification using activity attributes}
\label{sec:model:task}
We now want to classify composite activities that span an entire video sequence,
given attribute classifier scores. We note that we can use any of the scores introduced in the previous section ($s_{i,t}$, $s^{con}_{i,t}$, $s^{coocc}_{i,t}$ or their stacked combination). In the following for simplicity we refer to these scores as $s_{i,t}$ and corresponding matrix as $S$. In this approach we rely on the
representation that captures likelihoods of the presence or absence of a
particular attribute and leave modeling the temporal ordering of  attributes for
 future work.
We define a feature for the video sequence as $g^{seq}:\mathbb{R}^{n\times T}  \mapsto  \mathbb{R}^{n}$ by max pooling the scores of each attribute over all time intervals (see
\figref{fig:seqClassification}):
\begin{equation}
\label{eq:dishmax}
g^{seq}(S)=\max_{t\in\{1,...,T\}}s_{t} \end{equation}
\noindent where $\max$ is  an element-wise operator over all columns $s_t \in \mathbb{R}^n$ of matrix $S$.

To decide on the class $z$ of a sequence $d$ we use the feature $g^{seq}$ and classify it using a nearest neighbor classifier (\emph{NN}) or a one-versus-all SVM given a set of labeled training sequences. The SVM classifier provides us with the following confidence function for all composite classes $z$: $f^{seq}_z:\mathbb{R}^{n} \mapsto \mathbb{R}$, where the final score is defined as:
\begin{equation}
s^{seq}_{z,d} = f^{seq}_z(g^{seq}(S_d)),
\end{equation}
where $S_d$ is the score matrix for sequence $d$.
 The following sections describe alternatives to NN and SVM to incorporate prior knowledge mined from script data.

\subsection{Script data for recognizing composite activities}
\label{sec:model:semrel}
Composite activities show a high diversity which is practically impossible 
to capture in a training corpus. Our system thus needs to be robust against many activity variants that are not present in the training data. 
The use of attributes allows to include external knowledge to determine relevant attributes for a given composite activity. For this we assume associations between attribute $a_i$
and composite activity class $z$ in a matrix of weights $w_{z,i}$, with $Z$ being the number of composite activity classes. The vectors $w_z$ are L1 normalized, \ie $\sum_{i=1}^n w_{z,i}=1$.
Our system extracts those associations from script data (see \secref{sec:semrel}),
but the approach generalizes to other arbitrary external knowledge sources.
We explore three options to use such information which we detail in the following.
   \paragraph{Script data:} 
We compute the confidence $f^{scriptdata}_z:\mathbb{R}^{n}  \mapsto  \mathbb{R}$ of a sequence being of the composite activity  $z$ using the attribute-based feature representation $g^{seq}(S)$ introduced in \eqnref{eq:dishmax}. Given the  weights $w_{z,i}$ we compute a weighted sum: 
\begin{equation} 
f^{scriptdata}_z(g^{seq}(S)) =\sum_{i=1}^n w_{z,i} g^{seq}_i(S).
\label{eq:zeroshot}
\end{equation}
For a specific sequence $d$ with corresponding score matrix $S_d$ we get the following score:
\begin{equation} 
s^{scriptdata}_{z,d} = f^{scriptdata}_z(g^{seq}(S_d)).
\end{equation}

This formulation is similar to the sum formulation we used in \citep{rohrbach11cvpr} for image recognition with attributes, which itself is an adaption of the direct attribute prediction model introduced by \citet{lampert13pami}. Note that the weight matrix retrieved from script data is sparse (most $w_{z,i} = 0$). When mining from other corpora one might need to threshold the weights $w_{z,i}$, setting all others to zero, to achieve good performance as done \eg in \citep{rohrbach11cvpr}.

\paragraph{NN+script data:} When training data is available we can use a
nearest neighbor classifier. Often, only a handful of attributes
are likely to be indicative for a composite activity class, while the
majority of other attributes will provide irrelevant, potentially noisy information. When searching
for nearest neighbors such irrelevant attributes might dominate the distance,
resulting in suboptimal performance. %
To
reduce this effect we rely on the script data to constrain the attribute feature
vector to the relevant dimensions.

More specifically, we replace the L2 norm for computing the distance of nearest neighbor with the following training class dependent weighted L2 norm. It  takes weights of class-attribute associations into account. It is defined between the test attribute vector of unseen class $g^{seq}(S_{test})$ and the training attribute vector $g^{seq}(S_{train}^z)$ of class $z$ as: 

 \begin{multline}
Dist(S_{test},S_{train}^z) \\ =  \left(\sum_{i=1}^n w_{z,i}  \left(g^{seq}_i(S_{test})-g^{seq}_i(S_{train}^z) \right)^2\right)^{0.5}.
\label{eq:nn_script}
\end{multline}
To enhance robustness further, we binarize all association weights $w_{z,i}$ by setting
all non-zero weights to $1$ (and L1-normalize $w_z$). This reduces the distance computation to the
relevant attributes, normalized by the total number of relevant attributes. 

\paragraph{Propagated semantic transfer (PST):}
As the third approach to integrate external knowledge from script data we use Propagated semantic transfer (PST) which we proposed in \citep{rohrbach13nips} and summarize shortly in the following. The approach builds on \eqnref{eq:zeroshot} and uses label propagation to exploit the distances within the unlabeled data, \ie it assumes a transductive setting where all test data is available when predicting a single test label.

We can incorporate (partially) labeled training data $l_{z,d}\in \{0,1,\emptyset\}$ for class $z$ and sequence $d$. $\emptyset$ denotes that we do not have a label for this sequence and class. We combine the labels with the predictions in the following way, using only the most reliable predictions $s^{scriptdata}_{z,d}$ (top-$\delta$ fraction) per class $z$:
\begin{equation}
\label{eq:LabelCombi}
	s^{PST}_{z,d} = \begin{cases} \gamma{l}_{z,d}  & \text{if }{l}_{z,d} \in \{0,1\}  \\
	         	(1-\gamma) s^{scriptdata}_{z,d}   & \text{if among top-$\delta$ fraction} \\
	         	& \text{of predictions for class }z\\
	         	0  & \text{otherwise.}\\
	         \end{cases}
\end{equation}
$\gamma$ provides a weighting between the true labels and the predicted labels. In the zero-shot case we only use predictions and $\gamma = 0$. The parameters $\delta,\gamma\in [0,1]$ are chosen, similar to the remaining parameters, on the validation set.
For zero-shot we use the unlabeled training data as additional data for label propagation.

For computing the distance between the sequences we use the feature representation $g^{seq}(S)$, as for the \emph{NN}-classifier, which is much lower dimensional than the raw video feature representation and provides more reliable distances as we showed in \citep{rohrbach13nips}.
We build a k-NN graph by connecting the k closest neighbours. We set the weights of the graph edges between sequences $d$ and $e$ to $exp( -0.5 \sigma^{0.5}\|g^{seq}(S_d) - g^{seq}(S_e)\|)$, where $\sigma$ is set to the mean of the distances to the nearest neighbours.
We initialize this graph with the scores $s^{PST}_{z,d}$ and propagate them using label propagation from \citet{Zhou2004}.

\subsection{Prior knowledge from script data}
\label{sec:semrel}
We want to quantify what activities and objects typically occur in a composite activity by leveraging the script data we collected (see \secref{sec:mining-semrel}).
In order to use prior knowledge from textual script data, we have to match the (controlled) attribute labels from the video annotations to the (freely) written script instances (\secref{sec:labelMatching}). Based on the matched attributes we compute two different word frequency statistics (\secref{sec:semrel-stats}). 

\subsubsection{Label matching}
\label{sec:labelMatching}
To 
transfer any kind of knowledge from the script corpus to the attributes in the video annotation, we need to match attribute labels to natural language descriptions. %
The annotated attribute labels are standard English verbs (for activities, \emph{wash}) and nouns (for participating objects, \emph{carrot}), sometimes 
with additional particles (\emph{take apart} and \emph{take out}).
As the script instances contain freely written natural language sentences, they do not necessarily have any correspondence with the attribute label annotations. We compare two strategies for mapping annotations to script data sentences:
\begin{itemize}
	\item \textbf{literal}: we look for the exact matching of the attribute label within the data. 
		\item \textbf{WordNet}: we look for attribute labels and their synonyms. We take synonyms as members of the same \emph{synset} according to the WordNet ontology \citep{fellbaum:wordnet} and restrict them to words with the same part of speech, \ie we match only verbal synonyms to activity predicates and only nouns to object terms.
	\end{itemize}

\subsubsection{Statistics computed on the script data}
 \label{sec:semrel-stats}
We compute two different association scores between attribute labels $a_i$ and composite activities $z$. For this we concatenate all scripts for a given composite $z$ to a single document $\delta_z$.
\begin{itemize}
\item \textbf{freq}: word frequency $freq(a_i,\delta_z)$  for each attribute $a_i$ and composite activities  $z$.

\item \textbf{\tfidf} \citep[term frequency $*$ inverse document frequency,][]{Salton88term-weightingapproaches}  is a measure used in Information Retrieval to determine the relevance of a word for a document. Given a document collection $D=\{\delta_1,...,\delta_z,...,\delta_m\}$, \tfidf for a term or attribute $a_i$ and a document $\delta_z$  is computed as follows:
\begin{equation}
tfidf(a_i,\delta_z) = freq(a_i,\delta_z) * log\frac{|D|}{|\{\delta\in D:a_i \in \delta\}|},
\label{tfidf-scenario}
\end{equation}
where   $\{\delta\in D:a_i \in \delta\}$ is the set of documents containing $a_i$ at least once.  \tfidf represents the distinctiveness of a term for a document: the value increases if the term occurs often in the document and rarely in other documents.
\end{itemize}
We set $w_{z,i} =freq(a_i,\delta_z)$ or $w_{z,i} = tfidf(a_i,\delta_z)$ and L1-normalize all vectors $w_{z}$. These weights $w_{z,i}$ are then used in \eqnsref{eq:zeroshot} and
\eqref{eq:nn_script} and subsequently also in our PST approach.

\subsection{Automatic temporal segmentation}
\label{sec:model:segmentation}
While we assume a segmented video during training time to learn attribute classifiers as described in \secref{sec:approach:finegrained:det}, we want to segment the video automatically at test time. To avoid noisy and small segments we follow the idea we presented in \citep{rohrbach14gcpr}, namely we employ agglomerative clustering. We start with uniform intervals of 60 frames and describe each interval with an attribute-classifier score vector. We combine neighbouring intervals based on the cosine similarity of their score vectors and stop when we reach a threshold (found on the validation set). We aim for a segmentation with granularity similar to original manual annotation. After this a separately trained visual background classifier removes irrelevant or noisy segments. In our experiments we show that this leads to composite recognition results, similar to using the ground truth intervals for the attributes.

\section{Evaluation}
\label{sec:eval}
In this section we evaluate our approaches to fine-grained and composite activity recognition. We start with the fine-grained activity classification and detection and compare three types of approaches described in \secref{sec:approach:finegrained}, namely pose-based, hand-centric and holistic approaches. Next we evaluate our approaches for composite activity recognition introduced in \secref{sec:approach}, evaluating our attributes enhanced with context and co-occurrence, the recognition of composite cooking activities using different levels of supervision, and the  zero-shot approach using script data.

\subsection{Experimental Setup}
\label{sec:experimentalSetup}
This section details our experimental setup. We will release evaluation code to  reproduce and compare with our results.
See Table \ref{tbl:dataset} for the information on our training/validation/test split. We estimate all hyper parameters on the validation set and then retrain the models on the training and validation set with the best parameters.

\subsubsection{Experimental setup fine-grained activity classification and detection}
In the fine-grained recognition task we want to distinguish \DBnActivities fine-grained activities and \DBnObjects participating objects (see \tableref{tbl:results:fine-grained-activities} for the lists of activities and objects).
To learn the visual classifiers we use the annotated ground truth intervals provided with the dataset. We train one-vs-all SVMs using mean SGD \citep{rohrbach11cvpr} with a $\chi^2$ kernel approximation~\citep{vedaldi10cvpr}.
For detection we use the midpoint hit criterion to decide on the correctness of a detection, \ie the midpoint of the detection has to be within the ground-truth. If a second detection fires for one ground-truth label, it is counted as false positive.  
In the following we report the mean over the average precision (AP) of each class. Combining features is achieved by stacking the bag-of-word histograms.

\subsubsection{Experimental setup composite activity recognition}
For localizing attributes within composite activities we rely on our automatic segmentation (\secref{sec:model:segmentation}). %
We aim to recognize 31 composite activities (see bold names in Table \ref{tbl:list:dishes}).

We distinguish two cases for training the attributes with respect to composites.
\begin{description}
  \item[\emph{Attribute training on all composites.}] We use all available 218 training+validation videos for training the attribute classifiers. See left half of \tablesref{tbl:results:activities}, \ref{tbl:results:dish}, and \ref{tbl:results:scripts}.
   \item[\emph{Attribute training on disjoint composites.}] We use all available videos apart from those showing the test composite categories (in total 92 videos). This means that attributes and composites are trained on disjoint sets of composite categories and thus also on disjoint sets of videos. This tests how well novel composite categories can be recognized without additional attribute labels. See right half of \tablesref{tbl:results:activities}, \ref{tbl:results:dish}, and \ref{tbl:results:scripts}.
\end{description}
Next, we have two cases for training the composites.
\begin{description}
	\item[\emph{With training data for composites.}] We train on the 126 training+validation videos whose category is in the set of the 31 test categories. Note that in case of \emph{Attribute training on all composites} the training videos are also part of the attribute training. See top part of \tableref{tbl:results:dish}.
	\item[\emph{No training data for composites.}] Here we do not rely on any training labels for the composite activities. See bottom part of \tableref{tbl:results:dish} and all of \tableref{tbl:results:scripts}. Combined with \emph{Attribute training on disjoint composites} this is zero-shot recognition.
\end{description}

\subsection{Fine-grained activity classification and detection}
\label{sec:results:fine-grained}

\newcommand{\midruleClsResults}{\cmidrule(lr){1-2}  \cmidrule(lr){3-4} \cmidrule(lr){5-5}}
\begin{table}
\begin{center}
\begin{tabular}{r@{\ }l@{}r@{\ \ }r r}
\toprule   
\multicolumn{2}{l}{Approach}&Activities&Objects&All\\
\midruleClsResults
\multicolumn{4}{l}{\textbf{Pose-based approaches}}\\
(1)& BM       & 18.9 & 13.8 & 15.7\\
(2)& FFT      & 19.0 & 16.2 & 17.2\\
(3)& Combined & 24.1 & 19.0 & 20.8 \\
\midruleClsResults
\multicolumn{4}{l}{\textbf{Hand-centric approaches}}\\
(4)& Hand-cSift      & 23.0 & 23.8 &23.5\\
(5)& Hand-Trajectories & 45.1 & 31.5 & 36.4 \\
(6)& Combined & 43.5 & 34.2 &  37.5\\
\midruleClsResults
\multicolumn{4}{l}{\textbf{Holistic approach}}\\
(7)& Dense Trajectories & 44.5 & 31.3 & 36.1\\
\midruleClsResults
\multicolumn{4}{l}{\textbf{Combinations}}\\
(8)& Dense Traj,BM,FFT  & 43.1 & 30.7 & 35.2 \\
(9)& Dense Traj,Hand-Traj & 52.2 & 37.7 & 42.9\\
(10)& Dense Traj,Hand-Traj,-cSift & 51.2 & 39.3 & 43.7\\
\bottomrule  \end{tabular}
\end{center}
\caption{Fine-grained activity and object classification results, mean AP in \% (see \secref{sec:results:fine-grained} for discussion).}
\label{tbl:results:fineGrained:cls}
\end{table}

\myparagraph{Activity classification}
We start with the classification results on fine-grained activities and their participants (\tableref{tbl:results:fineGrained:cls}).

The body model features on the joint tracks (BM) achieve a mean average precision (AP) of 18.9\% for activities and 13.8\% for objects.
Comparing this to the FFT features, we observe that FFT performs slightly better, improving over BM the AP by 0.1\% and 2.4\% respectively.
The combination of BM and FFT features (line 3 in \tableref{tbl:results:fineGrained:cls}) yields a significant improvement, reaching AP of 24.1\% for activities and 19.0\% for objects. 
We attribute this to the complementary information encoded in the features. While BM encodes among others velocity-histograms of the joint-tracks and statistics between tracks of different joints, FFT features encode FFT coefficients of individual joints.
Still, this is a relatively low performance. It can be explained, on one hand, by failures of the pose estimation method and, on the other hand, the pose-based features might not contain enough information to successfully distinguish the challenging fine-grained activities and participating objects. 
Next we look at the performance of our proposed hand-centric features. Color Sift features, densely sampled in the hand neighborhood, allow us to improve the object recognition AP to 23.8\% (Hand-cSift), indicating their better suitability in particular for recognizing objects. 
Dense Trajectories features computed around hands (denoted as Hand-Trajectories) reach 45.1\% and 31.5\% recognition AP for activities and objects, respectively. Combining both features leads to a small disimprovement for activities, however it helps to further improve the object recognition performance to 34.2\%.
Overall our hand-centric approach reaches the recognition AP of 37.5\% for activities and objects together. 
The state-of-the-art holistic approach of Dense Trajectories \citep{wang13ijcv} obtains 44.5\% and 31.3\% recognition AP for activities and objects. If compared to our hand-centric features, this is slightly below the Hand-Trajectories, which are restricted to the areas around hands. This supports our hypothesis that the most relevant information for recognizing our fine-grained activities is contained in the hand regions. 
We also consider several feature combinations (lines 8, 9, 10 in \tableref{tbl:results:fineGrained:cls}). Combining Dense Trajectories with the pose-based features does not improve the recognition performance. However, combining them with Hand-Tra\-jectories improves the activity recognition by 7.7\% and object recognition by 6.4\% (line 7 vs 9 in \tableref{tbl:results:fineGrained:cls}). Finally, adding the Hand-cSift features allows to reach the impressive 43.7\% recognition AP for activities and objects together.    

The detailed comparison of Dense Trajectories, Hand-Trajectories and the final feature-combination (line 10 in \tableref{tbl:results:fineGrained:cls}) can be found in \tableref{tbl:results:fine-grained-activities}. Hand-Trajectories loose to Dense Trajectories on activities that include ``coarser'' motion, \eg \emph{push down}, \emph{hang} or \emph{plug}, and corresponding objects such as \emph{hook} or \emph{teapot}. Note that Hand-Trajectories outperform the Dense Trajectories for 35 activity classes, while in the opposite direction this holds only 25 times (for objects, respectively 65 vs 43 times). This shows again that the hand-centric features consistently outperform the holistic features in both tasks. 
Some example cases where the hand-centric approach is significantly better, are such activities as \emph{rip open}, \emph{take apart}, and \emph{grate} and such objects as \emph{cauliflower}, \emph{oven}, and \emph{cup}. At the same time the final feature combination (line 10 in \tableref{tbl:results:fineGrained:cls}) consistently outperforms both aforementioned features in about 60\% of cases. We demonstrate some qualitative results comparing Dense Trajectories to the final feature combination in \tableref{tbl:fine-grained:qualitative}.
We also looked closer at the performance of other features. \eg the combined pose features (line 3 in \tableref{tbl:results:fineGrained:cls}) perform well on ``coarser'', full-body activities, such as \emph{throw in garbage}, \emph{take out}, \emph{move}, while rather poorly on more fine-grained activities. On the other hand the Hand-cSift features are good in recognizing objects with distinct shapes/colors, \eg \emph{pineapple}, \emph{carrot}, \emph{bowl}, etc.

\begin{table}
\begin{center}
\begin{tabular}{r@{\ }l@{}r@{\ \ }r r}
\toprule   
\multicolumn{2}{l}{Approach}&Activities&Objects&All\\
\midruleClsResults
\multicolumn{4}{l}{\textbf{Pose-based approaches}}\\
(1)& BM       &9.7 & 7.6 & 8.3 \\
(2)& FFT      & 10.5 & 8.7 & 9.3 \\
(3)& Combined &14.3 & 9.8 & 11.4 \\
\midruleClsResults
\multicolumn{4}{l}{\textbf{Hand-centric approaches}}\\
(4)& Hand-cSift      & 10.5 & 10.9 & 10.7 \\
(5)& Hand-Trajectories & 21.3 & 14.0 & 16.6 \\
(6)& Combined & 26.0 & 20.6 & 22.5 \\
\midruleClsResults
\multicolumn{4}{l}{\textbf{Holistic approach}}\\
(7)& Dense Trajectories &  29.5 & 21.5 & 24.4 \\
\midruleClsResults
\multicolumn{4}{l}{\textbf{Combinations}}\\
(8)& Dense Traj,BM,FFT& 30.7 & 21.5 & 24.8 \\
(9)& Dense Traj,Hand-Traj &34.3 & 25.2 & 28.5 \\
(10)& Dense Traj,Hand-Traj,-cSift & 34.5 & 25.3 & 28.6 \\
\bottomrule  \end{tabular}
\end{center}
\caption{Fine-grained activity and object detection results, mean AP in \% (see \secref{sec:results:fine-grained} for discussion)}
\label{tbl:results:fineGrained:det}
\end{table}

\begin{table*}[!hp]
\begin{center}
\scriptsize
\renewcommand{\tabcolsep}{0.14cm}
\begin{tabular}{l rrr @{\hspace{0.3cm}}ll@{\hspace{0.3cm}} rrr@{\hspace{0.3cm}} ll rrr}
\cmidrule(lr){1-4} \cmidrule(lr){5-9}  \cmidrule(lr){10-14}
 & Dense & Hand &Combi & &  & Dense & Hand  & Combi  & &  & Dense  & Hand  & Combi  \\
 Activity& Traj & Traj & +cSift&&Object&Traj & Traj & +cSift&&Object&Traj & Traj & +cSift\\
\cmidrule(lr){1-4} \cmidrule(lr){5-9}  \cmidrule(lr){10-14} 
add & 19.8 & 16.3 & 24.0 & & apple & - & - & - & & mango & 3.8 & 7.0 & 2.5 \\ 
arrange & 61.9 & 32.1 & 33.8 & & arils & 19.8 & 57.8 & 12.5 & & masher & - & - & - \\ 
change temperature & 69.1 & 78.1 & 75.4 & & asparagus & - & - & - & & measuring-pitcher & 0.7 & 5.0 & 5.3 \\ 
chop & 36.6 & 35.4 & 48.3 & & avocado & 2.5 & 4.3 & 3.8 & & measuring-spoon & 34.1 & 12.6 & 7.3 \\ 
clean & 32.0 & 33.0 & 33.3 & & bag & - & - & - & & milk & 0.4 & 0.4 & 0.4 \\ 
close & 76.3 & 68.8 & 77.0 & & baking-paper & - & - & - & & mortar & - & - & - \\ 
cut apart & 33.8 & 36.2 & 33.5 & & baking-tray & - & - & - & & mushroom & - & - & - \\ 
cut dice & 39.3 & 45.7 & 44.9 & & blender & - & - & - & & net-bag & 0.3 & 0.2 & 0.7 \\ 
cut off ends & 21.4 & 52.0 & 31.9 & & bottle & 57.1 & 49.3 & 57.7 & & oil & 52.3 & 47.6 & 55.6 \\ 
cut out inside & 2.2 & 0.8 & 2.0 & & bowl & 34.7 & 33.1 & 49.0 & & onion & 19.3 & 20.4 & 22.7 \\ 
cut stripes & 12.9 & 13.0 & 15.4 & & box-grater & - & - & - & & orange & 18.4 & 11.1 & 19.3 \\ 
cut & 28.3 & 44.9 & 27.2 & & bread & 3.7 & 6.5 & 8.9 & & oregano & - & - & - \\ 
dry & 81.9 & 85.1 & 84.5 & & bread-knife & 3.0 & 4.0 & 8.1 & & oven & 30.7 & 73.4 & 89.3 \\ 
enter & 100.0 & 100.0 & 100.0 & & broccoli & 2.0 & 2.3 & 5.7 & & paper & - & - & - \\ 
fill & 94.3 & 90.8 & 86.2 & & bun & 1.2 & 2.3 & 8.5 & & paper-bag & 20.5 & 10.3 & 33.0 \\ 
gather & 25.7 & 23.8 & 35.7 & & bundle & 0.5 & 1.1 & 1.4 & & paper-box & 1.0 & 1.2 & 3.6 \\ 
grate & 66.7 & 100.0 & 100.0 & & butter & 6.2 & 1.9 & 9.6 & & parsley & 23.4 & 25.5 & 49.6 \\ 
hang & 85.8 & 57.2 & 81.4 & & carafe & 44.4 & 46.7 & 54.4 & & pasta & 26.1 & 16.0 & 40.7 \\ 
mix & 10.3 & 5.4 & 52.9 & & carrot & 26.5 & 41.3 & 64.9 & & peach & - & - & - \\ 
move & 75.7 & 75.7 & 78.3 & & cauliflower & 29.3 & 68.9 & 73.8 & & pear & - & - & - \\ 
open close & 60.8 & 65.7 & 64.7 & & cheese & - & - & - & & peel & 40.3 & 28.6 & 35.2 \\ 
open egg & 50.0 & 28.1 & 39.2 & & chefs-knife & 59.9 & 73.3 & 63.1 & & pepper & 3.1 & 14.4 & 6.7 \\ 
open tin & - & - & - & & chili & 0.6 & 0.9 & 1.3 & & peppercorn & - & - & - \\ 
open & 22.0 & 22.0 & 34.5 & & chive & - & - & - & & pestle & - & - & - \\ 
package & 0.4 & 1.6 & 1.8 & & chocolate & - & - & - & & philadelphia & - & - & - \\ 
peel & 55.0 & 67.2 & 58.6 & & coffee & 3.3 & 25.0 & 100.0 & & pineapple & 19.5 & 47.0 & 49.7 \\ 
plug & 41.6 & 32.6 & 81.0 & & coffee-container & 34.6 & 24.8 & 73.4 & & plastic-bag & 36.4 & 37.7 & 43.6 \\ 
pour & 44.8 & 44.9 & 45.1 & & coffee-machine & 34.7 & 65.1 & 91.2 & & plastic-bottle & 4.7 & 2.8 & 9.1 \\ 
pull apart & 38.7 & 53.8 & 45.2 & & coffee-powder & 0.5 & 1.3 & 3.0 & & plastic-box & 2.6 & 9.0 & 5.3 \\ 
pull up & 79.2 & 21.7 & 75.6 & & colander & 63.4 & 62.2 & 77.9 & & plastic-paper-bag & 0.9 & 14.7 & 19.6 \\ 
pull & 1.3 & 9.1 & 1.2 & & cooking-spoon & - & - & - & & plate & 65.7 & 69.2 & 73.9 \\ 
puree & - & - & - & & corn & - & - & - & & plum & 0.7 & 2.5 & 1.3 \\ 
purge & 0.1 & 0.1 & 0.6 & & counter & 71.8 & 70.3 & 76.5 & & pomegranate & 5.1 & 0.8 & 2.3 \\ 
push down & 30.7 & 7.6 & 28.0 & & cream & 0.9 & 0.5 & 1.4 & & pot & 84.3 & 88.0 & 91.1 \\ 
put in & 55.5 & 50.8 & 58.0 & & cucumber & 4.3 & 5.2 & 4.1 & & potato & 0.4 & 0.4 & 0.6 \\ 
put lid & 87.3 & 85.3 & 90.0 & & cup & 27.0 & 26.7 & 43.6 & & puree & - & - & - \\ 
put on & 6.2 & 5.6 & 1.2 & & cupboard & 97.5 & 98.0 & 98.4 & & raspberries & - & - & - \\ 
read & 5.1 & 5.4 & 5.6 & & cutting-board & 84.4 & 85.4 & 88.9 & & salad & - & - & - \\ 
remove from package & 19.3 & 34.3 & 31.5 & & dough & - & - & - & & salami & - & - & - \\ 
rip open & 2.8 & 45.0 & 100.0 & & drawer & 98.2 & 98.4 & 98.5 & & salt & 59.8 & 48.7 & 64.1 \\ 
scratch off & 30.7 & 33.1 & 31.9 & & egg & 12.1 & 3.6 & 7.3 & & seed & - & - & - \\ 
screw close & 77.3 & 77.5 & 77.5 & & eggshell & 3.5 & 3.6 & 11.2 & & side-peeler & 50.0 & 11.7 & 37.8 \\ 
screw open & 78.7 & 69.4 & 79.2 & & electricity-column & 89.3 & 82.3 & 98.1 & & sink & 47.0 & 54.0 & 53.9 \\ 
shake & 73.0 & 75.7 & 77.3 & & electricity-plug & 74.3 & 70.6 & 87.7 & & soup & - & - & - \\ 
shape & - & - & - & & fig & 1.0 & 1.0 & 0.9 & & spatula & 72.9 & 76.2 & 78.2 \\ 
slice & 47.2 & 71.3 & 57.4 & & filter-basket & 1.3 & 3.4 & 13.1 & & spice & 19.1 & 13.3 & 12.4 \\ 
smell & 49.7 & 15.7 & 33.0 & & finger & 18.4 & 15.4 & 8.8 & & spice-holder & 95.6 & 94.4 & 96.3 \\ 
spice & 88.6 & 89.0 & 89.2 & & flat-grater & 31.7 & 27.7 & 40.9 & & spice-shaker & 88.3 & 87.3 & 91.5 \\ 
spread & 87.1 & 77.1 & 96.7 & & flower-pot & - & - & - & & spinach & - & - & - \\ 
squeeze & 90.1 & 92.9 & 91.9 & & food & - & - & - & & sponge & 17.2 & 45.4 & 38.2 \\ 
stamp & - & - & - & & fork & 8.7 & 7.5 & 10.5 & & sponge-cloth & 67.1 & 68.1 & 75.0 \\ 
stir & 91.2 & 81.9 & 91.7 & & fridge & 100.0 & 99.8 & 100.0 & & spoon & 2.8 & 5.9 & 8.9 \\ 
strew & 1.7 & 2.4 & 2.4 & & front-peeler & 21.8 & 6.0 & 17.6 & & squeezer & 52.5 & 67.0 & 59.3 \\ 
take apart & 1.6 & 32.1 & 53.3 & & frying-pan & 88.7 & 91.9 & 93.6 & & stone & 0.2 & 0.7 & 0.7 \\ 
take lid & 66.2 & 76.8 & 71.7 & & garbage & 13.7 & 17.9 & 27.5 & & stove & 84.4 & 87.2 & 90.4 \\ 
take out & 94.1 & 93.9 & 95.1 & & garlic-bulb & 0.3 & 0.6 & 0.8 & & sugar & 22.0 & 24.2 & 29.0 \\ 
tap & 3.3 & 4.2 & 6.2 & & garlic-clove & 11.7 & 3.6 & 9.3 & & table-knife & - & - & - \\ 
taste & 9.4 & 21.0 & 22.0 & & ginger & 1.9 & 3.3 & 3.6 & & tap & 70.2 & 71.8 & 79.1 \\ 
test temperature & 11.3 & 11.8 & 35.1 & & glass & 2.6 & 4.5 & 21.6 & & tea-egg & 37.2 & 28.7 & 36.1 \\ 
throw in garbage & 96.7 & 96.0 & 97.1 & & green-beans & 21.1 & 24.6 & 23.2 & & tea-herbs & 60.5 & 55.6 & 91.1 \\ 
turn off & 7.4 & 21.1 & 33.0 & & ham & - & - & - & & teapot & 46.4 & 6.7 & 69.1 \\ 
turn on & 27.8 & 30.6 & 48.5 & & hand & 95.9 & 95.2 & 96.4 & & teaspoon & 29.2 & 32.4 & 36.5 \\ 
turn over & - & - & - & & handle & 100.0 & 9.1 & 100.0 & & tin & - & - & - \\ 
unplug & 8.7 & 3.8 & 20.0 & & hook & 95.6 & 71.2 & 98.3 & & tin-opener & - & - & - \\ 
wash & 93.4 & 93.9 & 93.7 & & \multicolumn{1}{p{2cm}}{hot-chocolate-powder-bag}  & - & - & - & & tissue & - & - & - \\ 
whip & - & - & - & & hot-dog & 2.1 & 2.7 & 8.8 & & toaster & 1.3 & 8.1 & 6.7 \\ 
wring out & 3.3 & 4.5 & 5.3 & & jar & 5.4 & 14.2 & 17.8 & & tomato & - & - & - \\ 
 &  &  & & & ketchup & 2.0 & 3.1 & 19.6 & & tongs & - & - & - \\ 
 &  &  & & & kettle-power-base & 14.4 & 9.8 & 41.4 & & top & - & - & - \\ 
 &  &  & & & kiwi & 1.1 & 2.9 & 1.5 & & towel & 73.2 & 76.9 & 79.2 \\ 
 &  &  & & & knife & 69.6 & 83.5 & 76.8 & & tube & 1.0 & 9.5 & 10.2 \\ 
 &  &  & & & knife-sharpener & - & - & - & & water & 55.0 & 46.9 & 57.2 \\ 
 &  &  & & & kohlrabi & - & - & - & & water-kettle & 40.7 & 25.9 & 53.7 \\ 
 &  &  & & & ladle & - & - & - & & wire-whisk & - & - & - \\ 
 &  &  & & & leek & 10.6 & 19.5 & 17.6 & & wrapping-paper & 2.9 & 0.4 & 2.0 \\ 
 &  &  & & & lemon & - & - & - & & yolk & 0.5 & 0.5 & 0.3 \\ 
 &  &  & & & lid & 67.1 & 70.8 & 71.8 & & zucchini & - & - & - \\ 
 &  &  & & & lime & 14.2 & 3.7 & 14.6 & &  &  &  &  \\ 
\cmidrule(lr){1-4} \cmidrule(lr){5-9}  \cmidrule(lr){10-14}
\end{tabular}
\caption{Fine-grained activities and object classification performance of Dense Trajectories, Hand Trajectories, and their combination including Hand-cSift (line 10 in \tableref{tbl:results:fineGrained:cls}) for 67 fine-grained activities and 155 participating objects. AP in \%. ``-'' denotes that the category is not part of the test set and not evaluated.}
\label{tbl:results:fine-grained-activities}
\end{center}
\end{table*}

\myparagraph{Activity detection}
Next we look at the detection performance (\tableref{tbl:results:fineGrained:det}), which is inherently more challenging than the classification task. Here the BM features reach 8.3\% overall AP and FFT get 9.3\%. Their combination (line 3 in \tableref{tbl:results:fineGrained:det}) gets 11.4\% overall AP, while Hand-cSift only reaches 10.7\%. Hand-Trajectories alone get 16.6\% AP and combined with Hand-cSift they reach 22.5\%, while the Dense Trajectories get 24.4\% AP. As we can see for this task our hand-centric features perform worse than holistic and even pose-based features (line 3 vs 4 in \tableref{tbl:results:fineGrained:det}). We believe the reason for this is that for correct segmentation of the video into activity intervals we need more holistic information, which the hand-centric features cannot provide, while pose-based and holistic features can capture it better. 
Similarly, when combining Dense Trajectories with the pose-based features (line 8 in \tableref{tbl:results:fineGrained:det}) we observe a small improvement, supporting our hypothesis that pose indeed helps to capture the detection boundaries.
On the other hand, combining Dense Trajectories with our hand-centric features significantly improves the performance, in particular by 4.7\% for activities and by 3.7\% for objects (line 6 vs 9 in \tableref{tbl:results:fineGrained:det}). Combining the obtained features with the Hand-cSift further improves the results and we reach the 28.6\% overall AP. The improvement obtained after combining holistic and hand-centric features can be explained by the increased classification AP within the obtained intervals.
We thus conclude that for activity detection we require holistic information, which can come \eg from the human pose. Combining the holistic and hand-centric features is still beneficial and significantly improves the performance.

\subsection{Context and co-occurrence for fine-grained activities}\label{sec:eval:attribute}
 \renewcommand{\midruleClsResults}{\cmidrule(lr){1-1}  \cmidrule(lr){2-2} \cmidrule(lr){3-6}}

\newcommand{\compositeHeader}{ \toprule\multicolumn{1}{r}{Attribute training on:} & \multicolumn{2}{c}{All} & \multicolumn{2}{c}{Disjoint}\\
&  \multicolumn{2}{c}{Composites} &  \multicolumn{2}{c}{Composites}\\
\cmidrule(lr){1-1} \cmidrule(lr){2-3}  \cmidrule(lr){4-5}
& Dense & Combi& Dense & Combi\\ 
&  Traj & +cSift &  Traj & +cSift \\
\cmidrule(lr){1-1} \cmidrule(lr){2-3}  \cmidrule(lr){4-5}}
\begin{table}[t]
\begin{center}
\begin{tabular}{l c@{\ \ }c  c@{\ \ }c}
\compositeHeader
(1) Base ($s^{base}$)& 			   36.1&\newcommand{\rAA}{43.7}\rAA          &33.5&35.9 \\
(2) Context only ($s^{con}$)&  11.1&\newcommand{\rAconA}{12.6}\rAconA    &\ 6.8 &\ 8.1\\
(3) Base+Context & 		       37.8&\newcommand{\rAconA}{41.2}\rAconA    &28.3&32.3\\
(4) Co-occ. only ($s^{coocc}$)&38.1&\newcommand{\rAcooccA}{41.7}\rAcooccA&32.6&35.3\\
(5) Base+Co-occ.&              38.1&\newcommand{\rAcooccA}{41.4}\rAcooccA&32.7&35.2\\
(6) Base+Cont.+Co-occ.&      39.3&\newcommand{\rAallA}{41.5}\rAallA    &30.8&32.6\\
\bottomrule \\ \end{tabular}
\caption[Attribute recognition using context and co-occurrence]{Attribute recognition using context and co-occurrence, mean AP in \%. Combi+cSift refers to Dense Traj,Hand-Traj,-cSift, see \secref{sec:eval:attribute} for discussion.}
\label{tbl:results:activities}
\end{center}
\end{table} 
While so far we looked at individual fine-grained activities, we now evaluate the benefit from co-occurrence and context as introduced in \secref{model:contextCooccurence}.
\tableref{tbl:results:activities} provides the results for recognizing activities and their participants, modeled as attributes. %
We evaluate in two settings. The left two columns of \tableref{tbl:results:activities} show the results for training on all composites in training set, while the right two columns are trained only on composites absent in test set (Disjoint Composites), \ie the second is a more challenging problem, as there is less training data and the attributes are tested in a different context. 
The performance in the first line is equivalent to the results in \tableref{tbl:results:fineGrained:cls}.
The very left column shows results on Dense Trajectories. More specifically using only temporal context to recognize activity attributes performance drops from 36.1\% AP for the base classifier to
11.1\% AP. This is the expected result, because the context is similar for all activities of the same sequence and thus cannot discriminate attributes.
In contrast, when using co-occurrence only (line 4 in \tableref{tbl:results:activities}), the performance increases by 2.0\% compared to the base classifiers due to the high relatedness between the attributes, namely between activities and their participants.
Combining context and co-occurrence information with the base classifier gives 37.8\% and 38.1\%, respectively. A combination of all training modes achieves a performance of 39.3\% AP, improving the base classifier's result by 3.2\%. While results for Dense Trajectories are as expected \ie adding context and co-occurrence improves performance, the performance drops slightly for the (in general) better performing combined features (second column). However, although the attribute prediction performance drops, we found that for recognizing the composites, context and co-occurrence are still useful.

In the second setting, we restrict the training dataset to composites absent in the test set (right two columns of \tableref{tbl:results:activities}), requiring the activity attributes to transfer to different composite activities. When comparing the right two the left columns, we notice a significant performance drop for all classifiers and both features. This decrease can mainly be attributed to the strong reduction of training data to about one third. The base classifier performs best and co-occurrence variants slightly below. Variants including context lead to tremendous performance drops in all combinations because the activity context changes from training to test (having different composite activities).

\subsection{Composite cooking activity classification}
\label{sec:eval:composites}
After evaluating attribute recognition performance in \secref{sec:eval:attribute}, we now show the results for recognizing composites as introduced in \secref{sec:model:task}. From the different attribute combination variants we only use the combination of base, context, and co-occurrence (last line in \tableref{tbl:results:activities}). Although this is not always the best choice for recognizing attributes we found it to work better or similar to alternatives for composite recognition. 
The results are shown in \tableref{tbl:results:tasks}, which, similar to \tableref{tbl:results:activities}, shows results for training the attributes on all composites, on the left, and reduced attribute training on non-test composites on the right. 
In the top section of the table we use training data for the composite cooking activities. In the bottom section of the table we use {\emph no} training data for the composite cooking activities. This is enabled by the use of script data as motivated before.
Disregarding the first line which does not use attributes at all and the second line which uses ground truth intervals for attributes, all other lines are based on attributes computed on our automatic temporal segmentation, introduced in  \secref{sec:model:segmentation}.

Examining the results in \tableref{tbl:results:tasks} we make several interesting observations.
First, training composites on attributes of fine-grained activities and objects (line 3 in \tableref{tbl:results:tasks}) outperforms low-level features (line 1 in \tableref{tbl:results:tasks}), supporting our claim that for learning composite activities it is important to share information on an intermediate level of attributes.

\begin{table}[t]
\begin{center}
\begin{tabular}{l c@{\ \ }c  c@{\ \ }c}
\compositeHeader
\multicolumn{5}{l}{\textbf{With training data for composites}}\\
\multicolumn{5}{l}{\emph{Without attributes}}\\
\ \ \ (1) SVM &    39.8   & 41.1& - & - \\
\multicolumn{5}{l}{\emph{Attributes on gt intervals}}\\
\ \ \ (2) SVM & 43.6   & 52.3& 32.3    &  34.9 \\ %
\multicolumn{5}{l}{\emph{Attributes on automatic segmentation}}\\ 
\ \ \ (3) SVM &                  49.0   & 56.9& 35.7    &   34.8\\
\ \ \ (4) NN &                   42.1   & 43.3& 24.7   &  32.7 \\
\ \ \ (5) NN+\Scriptknowledge&   35.0   & 40.4& 18.0    &21.9        \\
\ \ \ (6) PST+\Scriptknowledge&  54.5   & 57.4& 32.2  &32.5       \\ 
\cmidrule(lr){1-1} \cmidrule(lr){2-3}  \cmidrule(lr){4-5}
\multicolumn{5}{l}{\textbf{No training data for composites }}\\
\multicolumn{5}{l}{\emph{Attributes on automatic segmentation}}\\
\ \ \ (7) \Scriptknowledge& 36.7        & 29.9 & 19.6 &21.9\\
\ \ \ (8) PST + \Scriptknowledge& 36.6    & 43.8 & 21.1&19.3 \\
\bottomrule \\ \end{tabular}
\caption[Composite cooking activity classification.]{Composite cooking activity classification, mean AP in \%. Top left quarter: fully supervised, right column: reduced attribute training data, bottom section: no composite cooking activity training data, right bottom quarter: true zero shot. See \secref{sec:eval:composites} for discussion.}
\label{tbl:results:dish}
\label{tbl:results:tasks}
\end{center}
\end{table}

The second somewhat surprising observation is that recognizing composites based on our segmentation (line 3 in \tableref{tbl:results:tasks}) outperforms using ground truth segments (line 2 in \tableref{tbl:results:tasks}). We attribute this to the fact that our segmentation is coarser than the ground truth and that we additionally remove noisy and background segments with a background classifier. This leads to more robust attributes and consequently better composite recognition. 
This allows to have separate training sets for composites and attributes. This setting is explored in the top right quarter of \tableref{tbl:results:tasks}. Here the training sequences for attributes are disjoint with the ones for composites, \ie we do not require the attribute annotataions for the composite training set.

Third, the improvements we achieved for fine-grained activities and object recognition by combining hand-centric with holistic features are still evident for composites. The Combination of Dense Trajectoreis, Hand-Trajectories, and Hand-cSift (2\textsuperscript{nd}, 4\textsuperscript{th} column) outperforms in most cases Dense Trajectories only (1\textsuperscript{st}, 3\textsuperscript{rd} column), most notably in the setting ``All Composites'' for SVM (56.9\% over 49.0\% AP) and PST+\Scriptknowledge (43.8\% over 36.6\% AP). 

Fourth, using our Propagated Semantic Transfer (PST) approach is in most cases superior to other variants of incorporating script data (NN+\Scriptknowledge / \Scriptknowledge). Most notably it reaches 57.5\% AP for our combined feature. This is the overall best performance and also outperforms the SVM with 56.6\% AP. PST slightly drops for the last number in table (19.3\%), which we found is due to rather suboptimal parameters selected on the validations set. We note that in the scenario of Disjoint Composites (top right quarter of \tableref{tbl:results:tasks}) PST+\Scriptknowledge is outperformed by training an SVM. We attribute this to the fact that the attributes are  less robust in this scenario (see \tableref{tbl:results:activities}) and the SVM can better adjust to that by learning which attributes are reliable and which not. NN and PST are based on distances between attribute score vectors, thus metric learning could be beneficial in these cases.

Fifth, script data does not only allow to achieve the maximum performance but also allows transfer (bottom part of \tableref{tbl:results:tasks}) achieving in some cases results close to supervised approaches. The bottom right part of the table shows zero-shot recognition. Although here the performance cannot compete with the supervised setting, we like to point out that this is a very challenging scenario, where attributes are trained on different composites, without composite training data, and the video stream has to be segmented automatically.

\begin{table}[t]
\begin{center}
\begin{tabular}{l c@{\ \ }c  c@{\ \ }c}
\compositeHeader
\multicolumn{5}{l}{\textbf{No training data for composites }}\\
\Scriptknowledge\\
(1) freq-literal & 28.2 & 30.5 & 19.8 & 24.1\\
(2) freq-WN & 25.3 & 28.6 & 17.4 & 20.3\\
(3) \tfidf-literal & 35.9 & 31.8 & 20.0 & 23.6\\
(4) \tfidf-WN & 36.7 & 29.9 & 19.6 & 21.9\\
\bottomrule \\ 
\end{tabular}
\caption{Variants of script knowledge, AP in \%. Combi+cSift refers to Dense Traj,Hand-Traj,-cSift. See \secref{sec:eval:composites} for discussion.}
\label{tbl:results:scripts}
\end{center}
\end{table}

\begin{table*}[!p]
\begin{center}
\scriptsize
\begin{tabular}{p{1.45cm} p{3.0cm} p{3.0cm} p{3.0cm} p{3.0cm} p{1.5cm}}
& \includegraphics[scale=0.45]{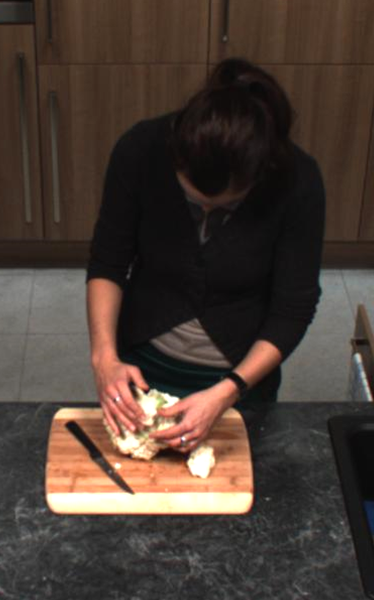} & \includegraphics[scale=0.45]{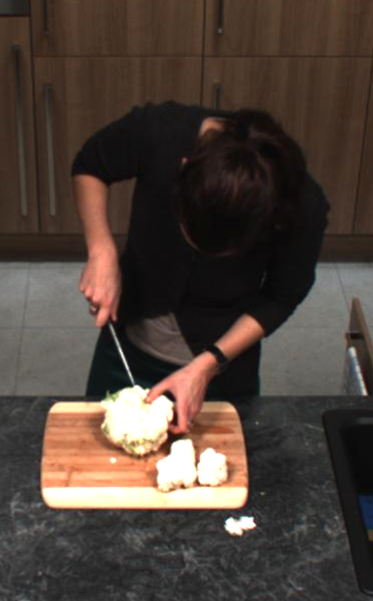} & \includegraphics[scale=0.45]{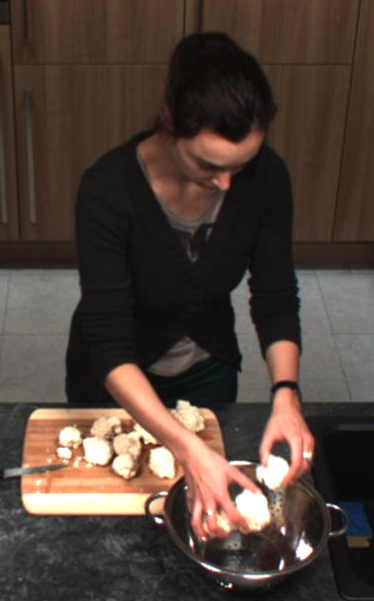} & \includegraphics[scale=0.45]{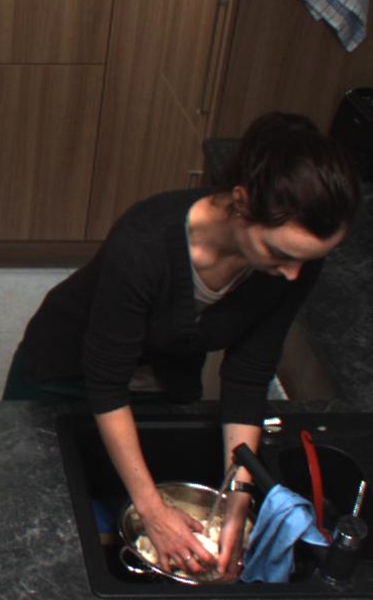} & Composites\\\cmidrule(lr){6-6}
Ground-truth & \textbf{cauliflower}, \textbf{cutting-board}, \textbf{hand}, \textbf{pull apart(A)} & \textbf{cauliflower}, \textbf{cut(A)}, \textbf{cutting-board}, \textbf{knife} & \textbf{add(A)}, \textbf{cauliflower}, \textbf{colander}, \textbf{cutting-board}, \textbf{hand} & \textbf{cauliflower}, \textbf{colander}, \textbf{hand}, \textbf{wash(A)} & \textbf{Preparing cauliflower}\vspace{2mm}\\
Dense Traj & \textbf{hand}, \textbf{cutting-board}, \textbf{pull apart(A)}, onion, peel, cut apart(A) & \textbf{knife}, \textbf{cutting-board}, cut apart(A), counter, chefs-knife, \textbf{cut(A)} & \textbf{hand}, \textbf{cutting-board}, move(A), counter, bowl, \textbf{colander} & \textbf{hand}, \textbf{wash(A)}, plate, \textbf{colander}, onion, peel &  Preparing orange\vspace{2mm}\\ 
Dense Traj, Hand-Traj, -cSift & \textbf{hand}, \textbf{cutting-board}, cut apart(A), \textbf{cauliflower}, onion, \textbf{pull apart(A)} & \textbf{cauliflower}, cut apart(A), \textbf{knife}, chefs-knife, \textbf{cutting-board}, \textbf{cut(A)} & \textbf{hand}, \textbf{cutting-board}, move(A), counter, \textbf{cauliflower}, \textbf{colander} & \textbf{hand}, \textbf{wash(A)}, bowl, \textbf{colander}, \textbf{cauliflower}, onion &  \textbf{Preparing cauliflower}\vspace{2mm}\\   
\cmidrule(lr){1-6}%
& \includegraphics[scale=0.45]{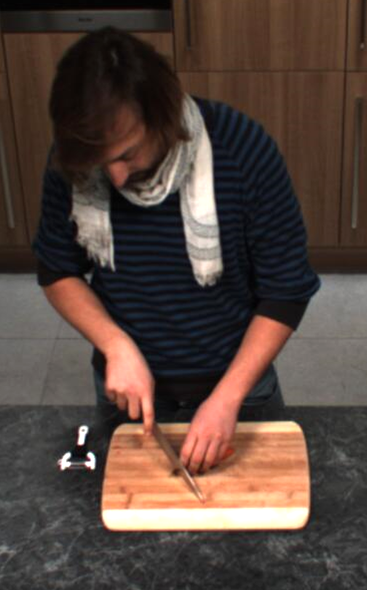} & \includegraphics[scale=0.45]{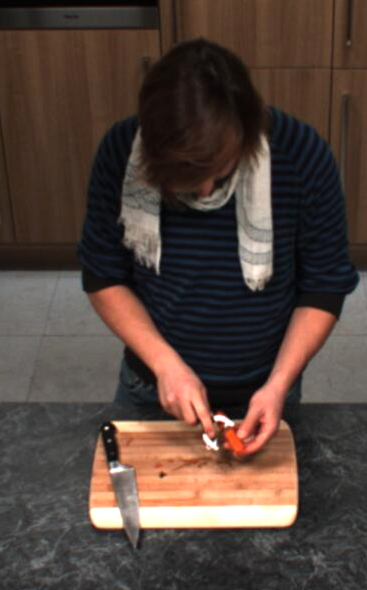} & \includegraphics[scale=0.45]{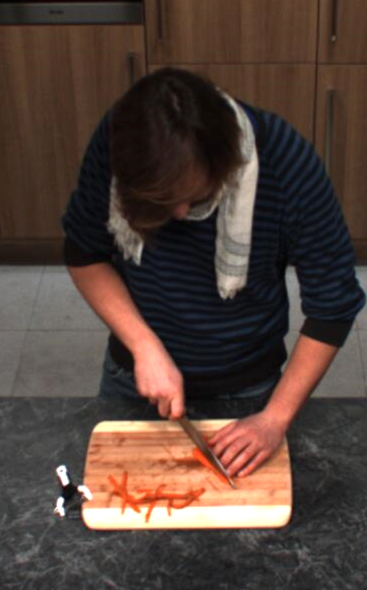} & \includegraphics[scale=0.45]{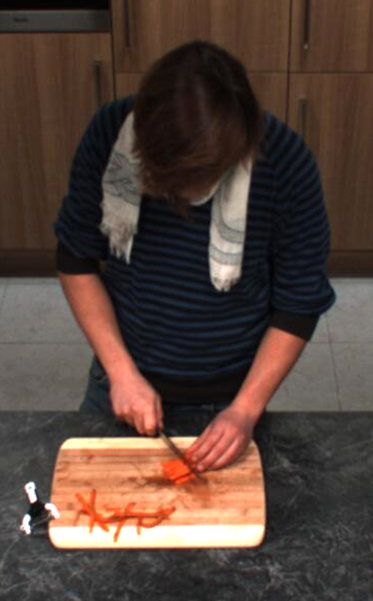} & Composites\\
\cmidrule(lr){6-6}
Ground-truth & \textbf{carrot}, \textbf{chefs-knife}, \textbf{cut off ends(A)}, \textbf{cutting-board} & \textbf{carrot}, \textbf{front-peeler}, \textbf{peel(A)} & \textbf{carrot}, \textbf{chefs-knife}, \textbf{cut stripes(A)}, \textbf{cutting-board} & \textbf{carrot}, \textbf{chefs-knife}, \textbf{cut apart(A)}, \textbf{cutting-board} & \textbf{Preparing carrot} \vspace{2mm}\\
Dense Traj & \textbf{cutting-board}, cut apart(A), \textbf{chefs-knife}, \textbf{cut off ends(A)}, knife, put on(A) & cutting-board, \textbf{peel(A)}, \textbf{front-peeler}, chefs-knife, knife, cucumber & \textbf{cutting-board}, \textbf{chefs-knife}, slice(A), knife, cut apart(A), cucumber & \textbf{cutting-board}, \textbf{cut apart(A)}, \textbf{chefs-knife}, knife, cauliflower, cut off ends(A) & Preparing cucumber\vspace{2mm}\\ 
Dense Traj, Hand-Traj, -cSift & \textbf{cutting-board}, \textbf{cut off ends(A)}, \textbf{chefs-knife}, cut apart(A), knife, \textbf{carrot} & cutting-board, \textbf{peel(A)}, \textbf{carrot}, chefs-knife, \textbf{front-peeler}, cucumber & \textbf{cutting-board}, \textbf{chefs-knife}, slice(A), knife, \textbf{carrot}, cut apart(A) & \textbf{cutting-board}, \textbf{cut apart(A)}, \textbf{chefs-knife}, cut off ends(A), knife, \textbf{carrot} & \textbf{Preparing carrot}\vspace{2mm}\\   
\cmidrule(lr){1-6}%
& \includegraphics[scale=0.45]{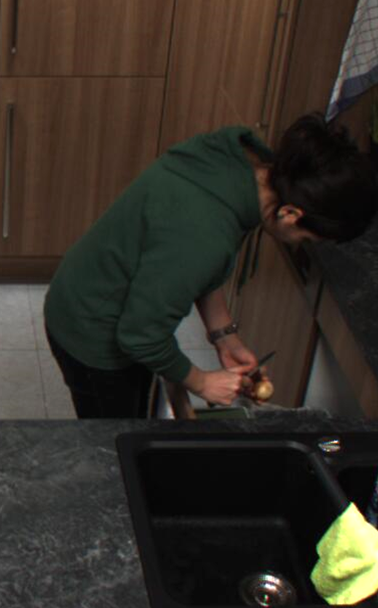} & \includegraphics[scale=0.45]{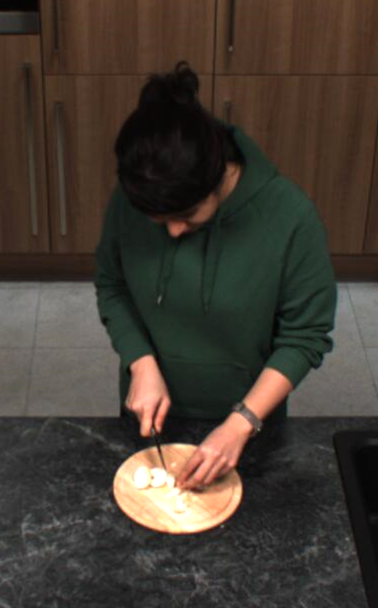} & \includegraphics[scale=0.45]{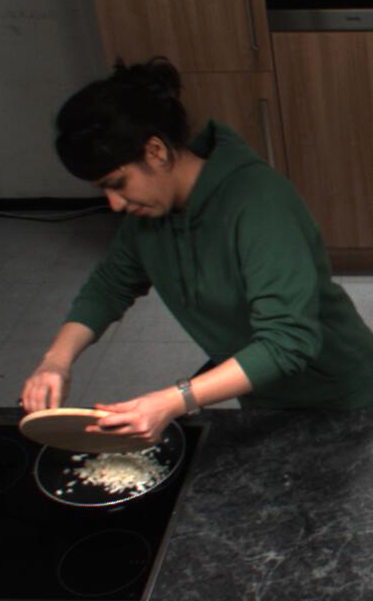} & \includegraphics[scale=0.45]{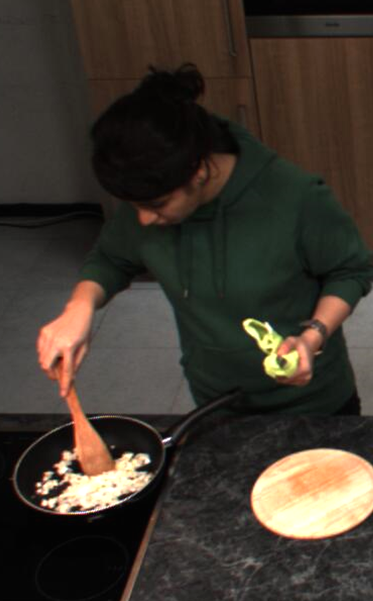} &  Composites\\
\cmidrule(lr){6-6}
Ground-truth & \textbf{knife}, \textbf{onion}, \textbf{peel(A)} & \textbf{chop(A)}, \textbf{cutting-board}, \textbf{knife,} \textbf{onion} & \textbf{add(A)}, \textbf{cutting-board}, \textbf{frying-pan}, \textbf{knife}, \textbf{onion} & \textbf{frying-pan}, \textbf{onion}, \textbf{spatula}, \textbf{stir(A)} & \textbf{Preparing onion}\vspace{2mm}\\
Dense Traj & \textbf{peel(A)}, hand, \textbf{onion}, throw in garbage(A), bowl, front-peeler & \textbf{cutting-board}, \textbf{knife}, cut dice(A), onion, \textbf{chop(A)}, slice(A) & hand, \textbf{frying-pan}, \textbf{cutting-board}, pot, spatula, \textbf{add(A)} & \textbf{spatula}, \textbf{frying-pan}, \textbf{stir(A)}, \textbf{onion}, add(A), egg &  \textbf{Preparing onion} \vspace{2mm}\\ 
Dense Traj, Hand-Traj, -cSift & \textbf{peel(A)}, hand, throw in garbage(A), \textbf{onion}, \textbf{knife}, peel & \textbf{cutting-board}, \textbf{knife}, cut dice(A), slice(A), \textbf{chop(A)}, chive & hand, \textbf{frying-pan}, \textbf{add(A)}, pot, spatula, cauliflower & \textbf{frying-pan}, \textbf{spatula}, \textbf{stir(A)}, \textbf{onion}, add(A), broccoli &  \textbf{Preparing onion} \vspace{2mm}\\   
\bottomrule 
\end{tabular}
\end{center}
\caption{Qualitative results for Dense Trajectories and its combination with hand-centric features (line 10 in \tableref{tbl:results:fineGrained:cls}) with respect to ground-truth. Top-6 highest scoring attributes (activities and objects) are shown, where (A) denotes activities. Composite activity predictions shown on the right. Correct results marked with bold. Note that many attributes are not correct according to the ground truth but very similar, \eg we predict \emph{slice} instead of \emph{cut stripes}.}
\label{tbl:fine-grained:qualitative}
\end{table*}

Sixth, while in \tableref{tbl:results:tasks} we always used the variant \tfidf-WN for \Scriptknowledge, we show different variants of \Scriptknowledge for the case where they are not combined with NN or PST in \tableref{tbl:results:scripts}. The main observation is that freq-WN performs in all cases worst, most likely the WordNet expansions make the results noisier.
While in the first column the \tfidf-WN works best, there is overall no clear winner. However, when incorporated in PST, it is more important to select appropriate parameters for PST on the validation set rather than selecting the right variant of \Scriptknowledge.

Last, we want to look at an interesting comparison of the first line (SVM without attributes) versus line 8 (PST + \Scriptknowledge), which effectively compares the settings ``only composite labels'' versus ``only attribute labels'' (+ \Scriptknowledge). Although the latter does not have any labels for the actual task of composite recognition it either performs close (in case of Dense Trajectories) or slightly better (for combined features). This indicates that our PST + \Scriptknowledge approach is very good in transferring information from the original task it was trained on to another which is very important for adaptation to novel situations, typical for assisted daily living scenarios. 

\tableref{tbl:fine-grained:qualitative} provides qualitative results for three composite videos including how they are decomposed into attributes of fine-grained activities and participating objects.

\section{Conclusion}
\label{sec:conlusion}

In this work we address two challenges that have not been widely explored so far, namely fine-grained activity recognition and composite activity recognition. In order to approach these tasks we propose the large activity database \MpiNew.
We recorded and annotated \DBnVideoSeq videos of more than \DBhours hours with \DBnSubjects human subjects performing a large number of realistic cooking activities. Our database is unique with respect to size, length, complexity of the videos, and available annotations (activities, objects, human pose, text descriptions). 

To estimate the complexity of fine-grained activity recognition in our database we compare three types of approaches: pose-based, hand-centric, and holistic. We evaluate on a classification and the often neglected detection task. Our results show that for recognizing fine-grained activities and their participating objects it is beneficial to focus on hand regions as the activities are hand-centric and the relevant objects are in the hand neighbourhood. 

Composite activities are difficult to recognize because of their inherent variability and the lack of training data for specific composites. We show that attribute-based activity recognition allows recognizing composite activities well. Most notably, we describe how textual script data, which is easy to collect, enables an improvement of the composite activity recognition when only little training data is available, and even allows for complete zero-shot transfer. 

As part of future work we plan to validate our hand-centric approach in other domains and exploit the scripts for composite activity recognition by modeling the temporal structure of the video.

\section*{Acknowledgments}This work was supported by a fellowship within the FITweltweit-Program of the German Academic Exchange Service (DAAD), by the Cluster of Excellence ``Multimodal Computing and Interaction'' of the German Excellence Initiative and the Max Planck Center for Visual Computing and Communication.

\bibliographystyle{plainnat}

\bibliography{biblioLong,biblioMarcus,rohrbach}

\begin{thebibliography}{116}
\providecommand{\natexlab}[1]{#1}
\providecommand{\url}[1]{\texttt{#1}}
\expandafter\ifx\csname urlstyle\endcsname\relax
  \providecommand{\doi}[1]{doi: #1}\else
  \providecommand{\doi}{doi: \begingroup \urlstyle{rm}\Url}\fi

\bibitem[Amin et~al.(2013)Amin, Andriluka, Rohrbach, and Schiele]{amin13bmvc}
Sikandar Amin, Mykhaylo Andriluka, Marcus Rohrbach, and Bernt Schiele.
\newblock {Multi-view Pictorial Structures for 3D Human Pose Estimation}.
\newblock In \emph{Proceedings of the British Machine Vision Conference
  (BMVC)}. {BMVA Press}, 2013.

\bibitem[Andriluka et~al.(2009)Andriluka, Roth, and Schiele]{Andriluka:2009}
Mykhaylo Andriluka, Stefan Roth, and Bernt Schiele.
\newblock Pictorial structures revisited: People detection and articulated pose
  estimation.
\newblock In \emph{Proceedings of the IEEE Conference on Computer Vision and
  Pattern Recognition (CVPR)}, 2009.

\bibitem[Andriluka et~al.(2011)Andriluka, Roth, and Schiele]{andriluka11ijcv}
Mykhaylo Andriluka, Stefan Roth, and Bernt Schiele.
\newblock Discriminative appearance models for pictorial structures.
\newblock \emph{International Journal of Computer Vision (IJCV)}, 2011.

\bibitem[Aubert and Pri{\'e}(2007)]{aubert07mm}
Olivier Aubert and Yannick Pri{\'e}.
\newblock Advene: an open-source framework for integrating and visualising
  audiovisual metadata.
\newblock In \emph{MM}. ACM, 2007.

\bibitem[Baccouche et~al.(2011)Baccouche, Mamalet, Wolf, Garcia, and
  Baskurt]{baccouche11hbu}
Moez Baccouche, Franck Mamalet, Christian Wolf, Christophe Garcia, and Atilla
  Baskurt.
\newblock Sequential deep learning for human action recognition.
\newblock In \emph{Human Behavior Understanding}, pages 29--39. Springer, 2011.

\bibitem[Barr and Feigenbaum(1981)]{barr82book}
Avron Barr and Edward Feigenbaum.
\newblock \emph{The Handbook of Artificial Intelligence, Volume 1}.
\newblock William Kaufman Inc., Los Altos, CA, 1981.

\bibitem[Bloem et~al.(2012)Bloem, Regneri, and
  Thater]{bloem-regneri-thater:KONVENS}
Jelke Bloem, Michaela Regneri, and Stefan Thater.
\newblock Robust processing of noisy web-collected data.
\newblock In \emph{KONVENS}, 2012.

\bibitem[Bojanowski et~al.(2014)Bojanowski, Lajugie, Bach, Laptev, Ponce,
  Schmid, and Sivic]{bojanowski14eccv}
Piotr Bojanowski, R\'emi Lajugie, Francis Bach, Ivan Laptev, Jean Ponce,
  Cordelia Schmid, and Josef Sivic.
\newblock Weakly supervised action labeling in videos under ordering
  constraints.
\newblock In \emph{Proceedings of the European Conference on Computer Vision
  (ECCV)}, 2014.

\bibitem[Brendel and Todorovic(2011)]{brendel11iccv}
William Brendel and Sinisa Todorovic.
\newblock Learning spatiotemporal graphs of human activities.
\newblock In \emph{Proceedings of the IEEE International Conference on Computer
  Vision (ICCV)}, 2011.

\bibitem[Campbell and Bobick(1995)]{campbell95iccv}
Lee Campbell and Aaron Bobick.
\newblock Recognition of human body motion using phase space constraints.
\newblock In \emph{Proceedings of the IEEE International Conference on Computer
  Vision (ICCV)}, 1995.

\bibitem[Chakraborty et~al.(2011)Chakraborty, Holte, Moeslund, Gonzalez, and
  Roca]{chakraborty11iccv}
Bhaskar Chakraborty, Michael Holte, Thomas Moeslund, Jordi Gonzalez, and Xavier
  Roca.
\newblock A selective spatio-temporal interest point detector for human action
  recognition in complex scenes.
\newblock In \emph{Proceedings of the IEEE International Conference on Computer
  Vision (ICCV)}, 2011.

\bibitem[Chaquet et~al.(2013)Chaquet, Carmona, and
  Fern\'{a}ndez-Caballero]{chaquet13cviu}
Jose Chaquet, Enrique Carmona, and Antonio Fern\'{a}ndez-Caballero.
\newblock A survey of video datasets for human action and activity recognition.
\newblock \emph{Computer Vision and Image Understanding}, 117\penalty0
  (6):\penalty0 633 -- 659, 2013.

\bibitem[Chen and Dolan(2011)]{chen11acl}
David Chen and William Dolan.
\newblock Collecting highly parallel data for paraphrase evaluation.
\newblock In \emph{Proceedings of the Annual Meeting of the Association for
  Computational Linguistics (ACL)}, 2011.

\bibitem[Cherian et~al.(2014)Cherian, Mairal, Alahari, and
  Schmid]{cherian14cvpr}
Anoop Cherian, Julien Mairal, Karteek Alahari, and Cordelia Schmid.
\newblock {Mixing Body-Part Sequences for Human Pose Estimation}.
\newblock In \emph{Proceedings of the IEEE Conference on Computer Vision and
  Pattern Recognition (CVPR)}, 2014.

\bibitem[Dalal et~al.(2006)Dalal, Triggs, and Schmid]{dalal06eccv}
Navneet Dalal, Bill Triggs, and Cordelia Schmid.
\newblock Human detection using oriented histograms of flow and appearance.
\newblock In \emph{Proceedings of the European Conference on Computer Vision
  (ECCV)}, 2006.

\bibitem[Das et~al.(2013)Das, Xu, Doell, and Corso]{das13cvpr}
Pradipto Das, Chenliang Xu, Richard Doell, and Jason Corso.
\newblock Thousand frames in just a few words: Lingual description of videos
  through latent topics and sparse object stitching.
\newblock In \emph{Proceedings of the IEEE Conference on Computer Vision and
  Pattern Recognition (CVPR)}, 2013.

\bibitem[Divvala et~al.(2012)Divvala, Efros, and Hebert]{Divvala:2012:HIA}
Santosh Divvala, Alexei Efros, and Martial Hebert.
\newblock How important are 'deformable parts' in the deformable parts model?
\newblock In \emph{Proceedings of the European Conference on Computer Vision
  Workshops (ECCV Workshops)}, 2012.

\bibitem[Elhoseiny et~al.(2013)Elhoseiny, Saleh, and Elgammal]{elhoseiny13iccv}
Mohamed Elhoseiny, Babak Saleh, and Ahmed Elgammal.
\newblock Write a classifier: Zero-shot learning using purely textual
  descriptions.
\newblock In \emph{Proceedings of the IEEE International Conference on Computer
  Vision (ICCV)}, 2013.

\bibitem[Everingham et~al.(2011)Everingham, Van~Gool, Williams, Winn, and
  Zisserman]{everingham11pascal}
Marc Everingham, Luc Van~Gool, Christopher Williams, John Winn, and Andrew
  Zisserman.
\newblock The {PASCAL} action classification taster competition, 2011.

\bibitem[Farhadi et~al.(2010)Farhadi, Endres, and Hoiem]{farhadi10cvpr}
Ali Farhadi, Ian Endres, and Derek Hoiem.
\newblock Attribute-centric recognition for cross-category generalization.
\newblock In \emph{Proceedings of the IEEE Conference on Computer Vision and
  Pattern Recognition (CVPR)}, 2010.

\bibitem[Fathi et~al.(2011)Fathi, Farhadi, and Rehg]{fathi11iccv}
Alireza Fathi, Ali Farhadi, and James Rehg.
\newblock Understanding egocentric activities.
\newblock In \emph{Proceedings of the IEEE International Conference on Computer
  Vision (ICCV)}. IEEE, 2011.

\bibitem[Fellbaum(1998)]{fellbaum:wordnet}
Christiane Fellbaum.
\newblock \emph{WordNet: An Electronical Lexical Database}.
\newblock The MIT Press, 1998.

\bibitem[Felzenszwalb and Huttenlocher(2005)]{Felzenszwalb:2005:PSO}
Pedro Felzenszwalb and Daniel Huttenlocher.
\newblock Pictorial structures for object recognition.
\newblock \emph{International Journal of Computer Vision (IJCV)}, 2005.

\bibitem[Felzenszwalb et~al.(2010)Felzenszwalb, Girshick, McAllester, and
  Ramanan]{Felzenszwalb2010PAMI}
Pedro Felzenszwalb, Ross Girshick, David McAllester, and Deva Ramanan.
\newblock Object detection with discriminatively trained part-based models.
\newblock \emph{IEEE Transactions on Pattern Analysis and Machine Intelligence
  (TPAMI)}, 32, 2010.

\bibitem[Ferrari et~al.(2008)Ferrari, Marin, and Zisserman]{Ferrari:2008:PSS}
Vittorio Ferrari, Manuel Marin, and Andrew Zisserman.
\newblock Progressive search space reduction for human pose estimation.
\newblock In \emph{Proceedings of the IEEE Conference on Computer Vision and
  Pattern Recognition (CVPR)}, 2008.

\bibitem[Ferryman(2007)]{ferryman07pets}
James Ferryman, editor.
\newblock \emph{PETS}, 2007.

\bibitem[Fischler and Elschlager(1973)]{Fischler1973TC}
Martin Fischler and Robert Elschlager.
\newblock The representation and matching of pictorial structures.
\newblock \emph{IEEE Trans. Comput'73}, 1973.

\bibitem[Frome et~al.(2013)Frome, Corrado, Shlens, Bengio, Dean, Ranzato, and
  Mikolov]{frome13nips}
Andrea Frome, Greg Corrado, Jon Shlens, Samy Bengio, Jeffrey Dean, Marc'Aurelio
  Ranzato, and Tomas Mikolov.
\newblock Devise: A deep visual-semantic embedding model.
\newblock In \emph{Advances in Neural Information Processing Systems (NIPS)},
  2013.

\bibitem[Fu et~al.(2013)Fu, Hospedales, Xiang, and Gong]{fu13pami}
Yanwei Fu, Timothy Hospedales, Tao Xiang, and Shaogang Gong.
\newblock Learning multi-modal latent attributes.
\newblock \emph{IEEE Transactions on Pattern Analysis and Machine Intelligence
  (TPAMI)}, PP\penalty0 (99), 2013.

\bibitem[Gkioxari et~al.(2013)Gkioxari, Arbelaez, Bourdev, and
  Malik]{gkioxari13cvpr}
Georgia Gkioxari, Pablo Arbelaez, Lubomir Bourdev, and Jitendra Malik.
\newblock Articulated pose estimation using discriminative armlet classifiers.
\newblock In \emph{Proceedings of the IEEE Conference on Computer Vision and
  Pattern Recognition (CVPR)}, 2013.

\bibitem[Guadarrama et~al.(2013)Guadarrama, Krishnamoorthy, Malkarnenkar,
  Venugopalan, Mooney, Darrell, and Saenko]{guadarrama13iccv}
Sergio Guadarrama, Niveda Krishnamoorthy, Girish Malkarnenkar, Subhashini
  Venugopalan, Raymond Mooney, Trevor Darrell, and Kate Saenko.
\newblock Youtube2text: Recognizing and describing arbitrary activities using
  semantic hierarchies and zero-shoot recognition.
\newblock In \emph{Proceedings of the IEEE International Conference on Computer
  Vision (ICCV)}, 2013.

\bibitem[Gupta et~al.(2009)Gupta, Srinivasan, Shi, and Davis]{gupta09cvpr}
Abhinav Gupta, Praveen Srinivasan, Jianbo Shi, and Larry Davis.
\newblock Understanding videos, constructing plots learning a visually grounded
  storyline model from annotated videos.
\newblock In \emph{Proceedings of the IEEE Conference on Computer Vision and
  Pattern Recognition (CVPR)}, 2009.

\bibitem[Jhuang et~al.(2013)Jhuang, Gall, Zuffi, Schmid, and
  Black]{jhuang13iccv}
Hueihan Jhuang, Jurgen Gall, Silvia Zuffi, Cordelia Schmid, and Michael Black.
\newblock {Towards understanding action recognition}.
\newblock In \emph{Proceedings of the IEEE International Conference on Computer
  Vision (ICCV)}, Sydney, Australia, 2013. IEEE.

\bibitem[Ji et~al.(2013)Ji, Xu, Yang, and Yu]{ji13tpami}
Shuiwang Ji, Wei Xu, Ming Yang, and Kai Yu.
\newblock {3D convolutional neural networks for human action recognition}.
\newblock \emph{IEEE Transactions on Pattern Analysis and Machine Intelligence
  (TPAMI)}, 35\penalty0 (1):\penalty0 221--231, 2013.

\bibitem[Kantorov and Laptev(2014)]{kantorov14cvpr}
Vadim Kantorov and Ivan Laptev.
\newblock {Efficient feature extraction, encoding and classification for action
  recognition}.
\newblock In \emph{Proceedings of the IEEE Conference on Computer Vision and
  Pattern Recognition (CVPR)}, 2014.

\bibitem[Karlinsky et~al.(2010)Karlinsky, Dinerstein, and
  Ullman]{karlinsky10nips}
Leonid Karlinsky, Michael Dinerstein, and Shimon Ullman.
\newblock Using body-anchored priors for identifying actions in single images.
\newblock In \emph{Advances in Neural Information Processing Systems (NIPS)},
  2010.

\bibitem[Karpathy et~al.(2014)Karpathy, Toderici, Shetty, Leung, Sukthankar,
  and Fei-Fei]{karpathy14cvpr}
Andrej Karpathy, George Toderici, Sanketh Shetty, Thomas Leung, Rahul
  Sukthankar, and Li~Fei-Fei.
\newblock {Large-scale video classification with convolutional neural
  networks}.
\newblock In \emph{Proceedings of the IEEE Conference on Computer Vision and
  Pattern Recognition (CVPR)}, 2014.

\bibitem[Kliper-Gross et~al.(2012)Kliper-Gross, Hassner, and
  Wolf]{kliper12pami}
Orit Kliper-Gross, Tal Hassner, and Lior Wolf.
\newblock The action similarity labeling challenge.
\newblock \emph{IEEE Transactions on Pattern Analysis and Machine Intelligence
  (TPAMI)}, 34\penalty0 (3):\penalty0 615--621, 2012.

\bibitem[Kuehne et~al.(2011)Kuehne, Jhuang, Garrote, Poggio, and
  Serre]{kuehne11iccv}
Hildegard Kuehne, Hueihan Jhuang, Est�baliz Garrote, Tomaso Poggio, and
  Thomas Serre.
\newblock Hmdb: A large video database for human motion recognition.
\newblock In \emph{Proceedings of the IEEE International Conference on Computer
  Vision (ICCV)}, 2011.

\bibitem[la~Torre et~al.(2009)la~Torre, Hodgins, Montano, Valcarcel, Forcada,
  and Macey]{torre09tr}
Fernando~De la~Torre, Jessica Hodgins, Javier Montano, Sergio Valcarcel, Ricard
  Forcada, and Justin Macey.
\newblock Guide to the cmu multimodal activity database.
\newblock Technical Report CMU-RI-TR-08-22, Robotics Institute, 2009.

\bibitem[Lampert et~al.(2013)Lampert, Nickisch, and Harmeling]{lampert13pami}
Christoph Lampert, Hannes Nickisch, and Stefan Harmeling.
\newblock Attribute-based classification for zero-shot learning of object
  categories.
\newblock \emph{IEEE Transactions on Pattern Analysis and Machine Intelligence
  (TPAMI)}, PP\penalty0 (99), 2013.

\bibitem[Laptev(2005)]{laptev05ijcv}
Ivan Laptev.
\newblock On space-time interest points.
\newblock In \emph{International Journal of Computer Vision (IJCV)}, 2005.

\bibitem[Laptev and P{\'e}rez(2007)]{laptev07iccv}
Ivan Laptev and Patrick P{\'e}rez.
\newblock Retrieving actions in movies.
\newblock In \emph{Proceedings of the IEEE International Conference on Computer
  Vision (ICCV)}, 2007.

\bibitem[Laptev et~al.(2008)Laptev, Marszalek, Schmid, and
  Rozenfeld]{laptev08cvpr}
Ivan Laptev, Marcin Marszalek, Cordelia Schmid, and Benjamin Rozenfeld.
\newblock Learning realistic human actions from movies.
\newblock In \emph{Proceedings of the IEEE Conference on Computer Vision and
  Pattern Recognition (CVPR)}, 2008.

\bibitem[Le et~al.(2011)Le, Zou, Yeung, and Ng]{le11cvpr}
Quoc Le, Will Zou, Serena Yeung, and Andrew Ng.
\newblock Learning hierarchical invariant spatio-temporal features for action
  recognition with independent subspace analysis.
\newblock In \emph{Proceedings of the IEEE Conference on Computer Vision and
  Pattern Recognition (CVPR)}, pages 3361--3368. IEEE, 2011.

\bibitem[Li and Li(2007)]{li07iccv}
Li-Jia Li and Fei-Fei Li.
\newblock What, where and who? classifying events by scene and object
  recognition.
\newblock In \emph{Proceedings of the IEEE International Conference on Computer
  Vision (ICCV)}, pages 1--8. IEEE, 2007.

\bibitem[Liu et~al.(2012)Liu, McCloskey, and Liu]{liu12icpr}
Jingchen Liu, S.~McCloskey, and Yanxi Liu.
\newblock Training data recycling for multi-level learning.
\newblock In \emph{Pattern Recognition (ICPR), 2012 21st International
  Conference on}, pages 2314--2318, Nov 2012.

\bibitem[Liu et~al.(2009)Liu, Luo, and Shah]{liu09cvpr}
Jingen Liu, Jiebo Luo, and Mubarak Shah.
\newblock Recognizing realistic actions from videos 'in the wild'.
\newblock In \emph{Proceedings of the IEEE Conference on Computer Vision and
  Pattern Recognition (CVPR)}, 2009.

\bibitem[Liu et~al.(2011)Liu, Kuipers, and Savarese]{liu11cvpr}
Jingen Liu, Benjamin Kuipers, and Silvio Savarese.
\newblock Recognizing human actions by attributes.
\newblock In \emph{Proceedings of the IEEE Conference on Computer Vision and
  Pattern Recognition (CVPR)}, 2011.

\bibitem[Marszalek et~al.(2009)Marszalek, Laptev, and Schmid]{marszalek09cvpr}
Marcin Marszalek, Ivan Laptev, and Cordelia Schmid.
\newblock Actions in context.
\newblock In \emph{Proceedings of the IEEE Conference on Computer Vision and
  Pattern Recognition (CVPR)}, june 2009.

\bibitem[Messing et~al.(2009)Messing, Pal, and Kautz]{messing09iccv}
Ross Messing, Chris Pal, and Henry Kautz.
\newblock Activity recognition using the velocity histories of tracked
  keypoints.
\newblock In \emph{Proceedings of the IEEE International Conference on Computer
  Vision (ICCV)}, 2009.

\bibitem[Mittal et~al.(2011)Mittal, Zisserman, and Torr]{mittal11bmvc}
Arpit Mittal, Andrew Zisserman, and Philip Torr.
\newblock Hand detection using multiple proposals.
\newblock In \emph{Proceedings of the British Machine Vision Conference
  (BMVC)}, 2011.

\bibitem[Motwani and Mooney(2012)]{motwani12ecai}
Tanvi~S. Motwani and Raymond~J. Mooney.
\newblock Improving video activity recognition using object recognition and
  text mining.
\newblock In \emph{ECAI}, pages 600--605, August 2012.

\bibitem[Natarajan and Nevatia(2008)]{natarajan08cvpr}
Pradeep Natarajan and Ramakant Nevatia.
\newblock View and scale invariant action recognition using multiview
  shape-flow models.
\newblock In \emph{Proceedings of the IEEE Conference on Computer Vision and
  Pattern Recognition (CVPR)}, 2008.

\bibitem[Niebles et~al.(2010)Niebles, Chen, and Fei-Fei]{niebles10eccv}
Juan Niebles, Chih-Wei Chen, and Li~Fei-Fei.
\newblock Modeling temporal structure of decomposable motion segments for
  activity classification.
\newblock In \emph{Proceedings of the European Conference on Computer Vision
  (ECCV)}, 2010.

\bibitem[Nilsback and Zisserman(2008)]{nilsback08icvgip}
Maria-Elena Nilsback and Andrew Zisserman.
\newblock Automated flower classification over a large number of classes.
\newblock In \emph{ICVGIP}, pages 722--729. IEEE, 2008.

\bibitem[Oh et~al.(2011)Oh, Hoogs, Perera, Cuntoor, Chen, Lee, Mukherjee,
  Aggarwal, Lee, Davis, Swears, Wang, Ji, Reddy, Shah, Vondrick, Pirsiavash,
  Ramanan, Yuen, Torralba, Song, Fong, Roy-Chowdhury, and Desai]{oh11cvpr}
Sangmin Oh, Anthony Hoogs, Amitha Perera, Naresh Cuntoor, Chia-Chih Chen,
  Jong~Taek Lee, Saurajit Mukherjee, Jake Aggarwal, Hyungtae Lee, Larry Davis,
  Eran Swears, Xiaoyang Wang, Qiang Ji, Kishore~K. Reddy, Mubarak Shah, Carl
  Vondrick, Hamed Pirsiavash, Deva Ramanan, Jenny Yuen, Antonio Torralba,
  Bi~Song, Anesco Fong, Amit Roy-Chowdhury, and Mita Desai.
\newblock A large-scale benchmark dataset for event recognition in surveillance
  video.
\newblock In \emph{Proceedings of the IEEE Conference on Computer Vision and
  Pattern Recognition (CVPR)}, pages 3153--3160. IEEE, 2011.

\bibitem[Over et~al.(2012)Over, Awad, Michel, Fiscus, Sanders, Shaw, Smeaton,
  and Qu\'{e}enot]{over12tv}
Paul Over, George Awad, Martial Michel, Jonathan Fiscus, Greg Sanders, B~Shaw,
  Alan~F. Smeaton, and Georges Qu\'{e}enot.
\newblock Trecvid 2012 -- an overview of the goals, tasks, data, evaluation
  mechanisms and metrics.
\newblock In \emph{Proceedings of TRECVID 2012}. NIST, USA, 2012.

\bibitem[Packer et~al.(2012)Packer, Saenko, and Koller]{packer12cvpr}
Benjamin Packer, Kate Saenko, and Daphne Koller.
\newblock A combined pose, object, and feature model for action understanding.
\newblock In \emph{Proceedings of the IEEE Conference on Computer Vision and
  Pattern Recognition (CVPR)}, 2012.

\bibitem[Patron-Perez et~al.(2010)Patron-Perez, Marszalek, Zisserman, and
  Reid]{patron10bmvc}
Alonso Patron-Perez, Marcin Marszalek, Andrew Zisserman, and Ian~D. Reid.
\newblock High five: Recognising human interactions in {TV} shows.
\newblock In \emph{Proceedings of the British Machine Vision Conference
  (BMVC)}, 2010.

\bibitem[Pirsiavash and Ramanan(2012)]{pirsiavash12cvpr}
Hamed Pirsiavash and Deva Ramanan.
\newblock Detecting activities of daily living in first-person camera views.
\newblock In \emph{Proceedings of the IEEE Conference on Computer Vision and
  Pattern Recognition (CVPR)}. IEEE, 2012.

\bibitem[Pirsiavash and Ramanan(2014)]{pirsiavash14cvpr}
Hamed Pirsiavash and Deva Ramanan.
\newblock {Parsing videos of actions with segmental grammars}.
\newblock In \emph{Proceedings of the IEEE Conference on Computer Vision and
  Pattern Recognition (CVPR)}, 2014.

\bibitem[Ramanathan et~al.(2013)Ramanathan, Liang, and
  Fei-Fei]{ramanathan13iccv}
Vignesh Ramanathan, Percy Liang, and Li~Fei-Fei.
\newblock Video event understanding using natural language descriptions.
\newblock In \emph{Proceedings of the IEEE International Conference on Computer
  Vision (ICCV)}, 2013.

\bibitem[Raptis and Sigal(2013)]{raptis13cvpr}
Michalis Raptis and Leonid Sigal.
\newblock Poselet key-framing: A model for human activity recognition.
\newblock In \emph{Proceedings of the IEEE Conference on Computer Vision and
  Pattern Recognition (CVPR)}, 2013.

\bibitem[Regneri et~al.(2010)Regneri, Koller, and Pinkal]{regneri10acl}
Michaela Regneri, Alexander Koller, and Manfred Pinkal.
\newblock Learning script knowledge with web experiments.
\newblock In \emph{Proceedings of the Annual Meeting of the Association for
  Computational Linguistics (ACL)}, 2010.

\bibitem[Regneri et~al.(2013)Regneri, Rohrbach, Wetzel, Thater, Schiele, and
  Pinkal]{regneri13tacl}
Michaela Regneri, Marcus Rohrbach, Dominikus Wetzel, Stefan Thater, Bernt
  Schiele, and Manfred Pinkal.
\newblock {Grounding Action Descriptions in Videos}.
\newblock \emph{Transactions of the Association for Computational Linguistics
  (TACL)}, 1, 2013.

\bibitem[Rodriguez et~al.(2008)Rodriguez, Ahmed, and Shah]{rodriguez08cvpr}
Mikel Rodriguez, Javed Ahmed, and Mubarak Shah.
\newblock Action {MACH} a spatio-temporal maximum average correlation height
  filter for action recognition.
\newblock In \emph{Proceedings of the IEEE Conference on Computer Vision and
  Pattern Recognition (CVPR)}, 2008.

\bibitem[Roggen et~al.(2010)Roggen, Calatroni, Rossi, Holleczek, Forster,
  Troster, Lukowicz, Bannach, Pirkl, Ferscha, Doppler, Holzmann, Kurz, Holl,
  Chavarriaga, Sagha, Bayati, Creatura, and del R.~Millan]{roggen10icnss}
Daniel Roggen, Alberto Calatroni, Mirco Rossi, Thomas Holleczek, Kilian
  Forster, Gerhard Troster, Paul Lukowicz, David Bannach, Gerald Pirkl, Alois
  Ferscha, Jakob Doppler, Clemens Holzmann, Marc Kurz, Gerald Holl, Ricardo
  Chavarriaga, Hesam Sagha, Hamidreza Bayati, Marco Creatura, and Jose del
  R.~Millan.
\newblock Collecting complex activity data sets in highly rich networked sensor
  environments.
\newblock In \emph{INSS}, 2010.

\bibitem[Rohrbach et~al.(2014)Rohrbach, Rohrbach, Qiu, Friedrich, Pinkal, and
  Schiele]{rohrbach14gcpr}
Anna Rohrbach, Marcus Rohrbach, Wei Qiu, Annemarie Friedrich, Manfred Pinkal,
  and Bernt Schiele.
\newblock Coherent multi-sentence video description with variable level of
  detail.
\newblock In \emph{Proceedings of the German Confeence on Pattern Recognition
  (GCPR)}, September 2014.

\bibitem[Rohrbach et~al.(2015)Rohrbach, Rohrbach, Tandon, and
  Schiele]{rohrbach15cvpr}
Anna Rohrbach, Marcus Rohrbach, Niket Tandon, and Bernt Schiele.
\newblock A dataset for movie description.
\newblock In \emph{Proceedings of the IEEE Conference on Computer Vision and
  Pattern Recognition (CVPR)}, 2015.

\bibitem[Rohrbach et~al.(2010)Rohrbach, Stark, Szarvas, Gurevych, and
  Schiele]{rohrbach10cvpr}
Marcus Rohrbach, Michael Stark, Gy{\"o}rgy Szarvas, Iryna Gurevych, and Bernt
  Schiele.
\newblock {What helps Where - and Why? Semantic Relatedness for Knowledge
  Transfer}.
\newblock In \emph{Proceedings of the IEEE Conference on Computer Vision and
  Pattern Recognition (CVPR)}, 2010.

\bibitem[Rohrbach et~al.(2011)Rohrbach, Stark, and Schiele]{rohrbach11cvpr}
Marcus Rohrbach, Michael Stark, and Bernt Schiele.
\newblock {Evaluating Knowledge Transfer and Zero-Shot Learning in a
  Large-Scale Setting}.
\newblock In \emph{Proceedings of the IEEE Conference on Computer Vision and
  Pattern Recognition (CVPR)}, 2011.

\bibitem[Rohrbach et~al.(2012{\natexlab{a}})Rohrbach, Amin, Andriluka, and
  Schiele]{rohrbach12cvpr}
Marcus Rohrbach, Sikandar Amin, Mykhaylo Andriluka, and Bernt Schiele.
\newblock {A database for fine grained activity detection of cooking
  activities}.
\newblock In \emph{Proceedings of the IEEE Conference on Computer Vision and
  Pattern Recognition (CVPR)}, 2012{\natexlab{a}}.

\bibitem[Rohrbach et~al.(2012{\natexlab{b}})Rohrbach, Regneri, Andriluka, Amin,
  Pinkal, and Schiele]{rohrbach12eccv}
Marcus Rohrbach, Michaela Regneri, Mykhaylo Andriluka, Sikandar Amin, Manfred
  Pinkal, and Bernt Schiele.
\newblock {Script data for attribute-based recognition of composite
  activities}.
\newblock In \emph{Proceedings of the European Conference on Computer Vision
  (ECCV)}, 2012{\natexlab{b}}.

\bibitem[Rohrbach et~al.(2013{\natexlab{a}})Rohrbach, Ebert, and
  Schiele]{rohrbach13nips}
Marcus Rohrbach, Sandra Ebert, and Bernt Schiele.
\newblock {Transfer Learning in a Transductive Setting}.
\newblock In \emph{Advances in Neural Information Processing Systems (NIPS)},
  2013{\natexlab{a}}.

\bibitem[Rohrbach et~al.(2013{\natexlab{b}})Rohrbach, Qiu, Titov, Thater,
  Pinkal, and Schiele]{rohrbach13iccv}
Marcus Rohrbach, Wei Qiu, Ivan Titov, Stefan Thater, Manfred Pinkal, and Bernt
  Schiele.
\newblock Translating video content to natural language descriptions.
\newblock In \emph{Proceedings of the IEEE International Conference on Computer
  Vision (ICCV)}, 2013{\natexlab{b}}.

\bibitem[Ryoo and Aggarwal(2009)]{ryoo09iccv}
Michael Ryoo and Jake Aggarwal.
\newblock Spatio-temporal relationship match: Video structure comparison for
  recognition of complex human activities.
\newblock In \emph{Proceedings of the IEEE International Conference on Computer
  Vision (ICCV)}, 2009.

\bibitem[Salton and Buckley(1988)]{Salton88term-weightingapproaches}
Gerard Salton and Christopher Buckley.
\newblock Term-weighting approaches in automatic text retrieval.
\newblock In \emph{Information Processing And Management}, 1988.

\bibitem[Sapp et~al.(2010)Sapp, Toshev, and Taskar]{Sapp10cascadedmodels}
Benjamin Sapp, Alexander Toshev, and Ben Taskar.
\newblock Cascaded models for articulated pose estimation, 2010.

\bibitem[Schuldt et~al.(2004)Schuldt, Laptev, and Caputo]{schuldt04icpr}
Christian Schuldt, Ivan Laptev, and Barbara Caputo.
\newblock Recognizing human actions: a local {SVM} approach.
\newblock In \emph{ICPR}, 2004.

\bibitem[Senina et~al.(2014)Senina, Rohrbach, Qiu, Friedrich, Amin, Andriluka,
  Pinkal, and Schiele]{senina14arxiv}
Anna Senina, Marcus Rohrbach, Wei Qiu, Annemarie Friedrich, Sikandar Amin,
  Mykhaylo Andriluka, Manfred Pinkal, and Bernt Schiele.
\newblock Coherent multi-sentence video description with variable level of
  detail.
\newblock \emph{arXiv:1403.6173}, 03/2014 2014.

\bibitem[Shotton et~al.(2011)Shotton, Fitzgibbon, Cook, Sharp, Finocchio,
  Moore, Kipman, and Blake]{shotton11cvpr}
Jamie Shotton, Andrew Fitzgibbon, Mat Cook, Toby Sharp, Mark Finocchio, Richard
  Moore, Alex Kipman, and Andrew Blake.
\newblock Real-time human pose recognition in parts from single depth images.
\newblock In \emph{Proceedings of the IEEE Conference on Computer Vision and
  Pattern Recognition (CVPR)}, pages 1297--1304. IEEE, 2011.

\bibitem[Sill et~al.(2009)Sill, Tak{\'a}cs, Mackey, and Lin]{sill09arxiv}
Joseph Sill, G{\'a}bor Tak{\'a}cs, Lester Mackey, and David Lin.
\newblock Feature-weighted linear stacking.
\newblock \emph{arXiv:0911.0460}, 2009.

\bibitem[Singh et~al.(2002)Singh, Lin, Mueller, Lim, Perkins, and Zhu]{omcs2}
Push Singh, Thomas Lin, Erik Mueller, Grace Lim, Travell Perkins, and Wan Zhu.
\newblock Open mind common sense: Knowledge acquisition from the general
  public.
\newblock In \emph{DOA, CoopIS and ODBASE 2002}, 2002.

\bibitem[Singh and Nevatia(2011)]{singh11iccv}
Vivek Singh and Ram Nevatia.
\newblock Action recognition in cluttered dynamic scenes using pose-specific
  part models.
\newblock In \emph{Proceedings of the IEEE International Conference on Computer
  Vision (ICCV)}, 2011.

\bibitem[Socher and Fei-Fei(2010)]{socher10cvpr}
Richard Socher and Li~Fei-Fei.
\newblock Connecting modalities: Semi-supervised segmentation and annotation of
  images using unaligned text corpora.
\newblock In \emph{Proceedings of the IEEE Conference on Computer Vision and
  Pattern Recognition (CVPR)}, San Francisco, CA, June 2010.

\bibitem[Socher et~al.(2013)Socher, Ganjoo, Manning, and Ng]{socher13nips}
Richard Socher, Milind Ganjoo, Christopher~D. Manning, and Andrew Ng.
\newblock Zero-shot learning through cross-modal transfer.
\newblock In \emph{Advances in Neural Information Processing Systems (NIPS)},
  pages 935--943, 2013.

\bibitem[Soomro et~al.(2012)Soomro, Zamir, and Shah]{soomro12arxiv}
Khurram Soomro, Amir~Roshan Zamir, and Mubarak Shah.
\newblock Ucf101: A dataset of 101 human actions classes from videos in the
  wild.
\newblock Technical report, arXiv:1212.0402, 2012.

\bibitem[Srikantha and Gall(2014)]{srikantha14eccv}
Abhilash Srikantha and Juergen Gall.
\newblock Discovering object classes from activities.
\newblock In \emph{Proceedings of the European Conference on Computer Vision
  (ECCV)}, pages 415--430. Springer, 2014.

\bibitem[Stein and McKenna(2013)]{stein13acm}
Sebastian Stein and Stephen McKenna.
\newblock Combining embedded accelerometers with computer vision for
  recognizing food preparation activities.
\newblock In \emph{UbiComp}. ACM, September 2013.

\bibitem[Sung et~al.(2011)Sung, Ponce, Selman, and Saxena]{sung11corr}
Jaeyong Sung, Colin Ponce, Bart Selman, and Ashutosh Saxena.
\newblock Human activity detection from {RGBD} images.
\newblock \emph{CoRR}, abs/1107.0169, 2011.
\newblock informal publication.

\bibitem[Tang et~al.(2012)Tang, Fei-Fei, and Koller]{tang12cvpr}
Kevin Tang, Li~Fei-Fei, and Daphne Koller.
\newblock Learning latent temporal structure for complex event detection.
\newblock In \emph{Proceedings of the IEEE Conference on Computer Vision and
  Pattern Recognition (CVPR)}, Providence, RI, USA, June 2012.

\bibitem[Tang et~al.(2013)Tang, Yao, Fei-Fei, and Koller]{tang13iccv}
Kevin Tang, Bangpeng Yao, Li~Fei-Fei, and Daphne Koller.
\newblock Combining the right features for complex event recognition.
\newblock In \emph{Proceedings of the IEEE International Conference on Computer
  Vision (ICCV)}, 2013.

\bibitem[Taylor et~al.(2010)Taylor, Fergus, LeCun, and Bregler]{taylor10eccv}
Graham~W Taylor, Rob Fergus, Yann LeCun, and Christoph Bregler.
\newblock Convolutional learning of spatio-temporal features.
\newblock In \emph{Proceedings of the European Conference on Computer Vision
  (ECCV)}, pages 140--153. Springer, 2010.

\bibitem[Tenorth et~al.(2009)Tenorth, Bandouch, and Beetz]{tenorth09iccw}
Moritz Tenorth, Jan Bandouch, and Michael Beetz.
\newblock {The {TUM} Kitchen Data Set of Everyday Manipulation Activities for
  Motion Tracking and Action Recognition}.
\newblock In \emph{THEMIS}, 2009.

\bibitem[Teo et~al.(2012)Teo, Yang, Daume, Fermuller, and Aloimonos]{teo12icra}
Ching~Lik Teo, Yezhou Yang, H~Daume, C~Fermuller, and Yiannis Aloimonos.
\newblock Towards a watson that sees: Language-guided action recognition for
  robots.
\newblock In \emph{Proceedings of the IEEE International Conference on Robotics
  and Automation (ICRA)}, pages 374--381. IEEE, 2012.

\bibitem[Ting and Witten(1997)]{ting97stacked}
Kai~Ming Ting and Ian~H Witten.
\newblock Stacked generalization: when does it work?
\newblock In \emph{Proceedings of the International Joint Conference on
  Artificial Intelligence (IJCAI)}, 1997.

\bibitem[Vedaldi and Zisserman(2010)]{vedaldi10cvpr}
Andrea Vedaldi and Andrew Zisserman.
\newblock Efficient additive kernels via explicit feature maps.
\newblock In \emph{Proceedings of the IEEE Conference on Computer Vision and
  Pattern Recognition (CVPR)}, 2010.

\bibitem[Wang and Schmid(2013)]{wang13iccv}
Heng Wang and Cordelia Schmid.
\newblock Action recognition with improved trajectories.
\newblock In \emph{Proceedings of the IEEE International Conference on Computer
  Vision (ICCV)}, Sydney, Australia, 2013.

\bibitem[Wang et~al.(2009{\natexlab{a}})Wang, Ullah, Klaser, Laptev, and
  Schmid]{wang09bmvc}
Heng Wang, Muhammad Ullah, Alexander Klaser, Ivan Laptev, and Cordelia Schmid.
\newblock {Evaluation of local spatio-temporal features for action
  recognition}.
\newblock In \emph{Proceedings of the British Machine Vision Conference
  (BMVC)}, 2009{\natexlab{a}}.

\bibitem[Wang et~al.(2011)Wang, Kl{\"a}ser, Schmid, and Liu]{wang11cvpr}
Heng Wang, Alexander Kl{\"a}ser, Cordelia Schmid, and Cheng-Lin Liu.
\newblock {Action Recognition by Dense Trajectories}.
\newblock In \emph{Proceedings of the IEEE Conference on Computer Vision and
  Pattern Recognition (CVPR)}, 2011.

\bibitem[Wang et~al.(2013{\natexlab{a}})Wang, Kl{\"a}ser, Schmid, and
  Liu]{wang13ijcv}
Heng Wang, Alexander Kl{\"a}ser, Cordelia Schmid, and C.L. Liu.
\newblock Dense trajectories and motion boundary descriptors for action
  recognition.
\newblock \emph{International Journal of Computer Vision (IJCV)},
  2013{\natexlab{a}}.

\bibitem[Wang et~al.(2009{\natexlab{b}})Wang, Markert, and
  Everingham]{wangME09bmvc}
Josiah Wang, Katja Markert, and Mark Everingham.
\newblock Learning models for object recognition from natural language
  descriptions.
\newblock In Andrea Cavallaro, Simon Prince, and Daniel~C. Alexander, editors,
  \emph{Proceedings of the British Machine Vision Conference (BMVC)}, pages
  1--11. British Machine Vision Association, 2009{\natexlab{b}}.

\bibitem[Wang et~al.(2013{\natexlab{b}})Wang, Qiao, and Tang]{wangYuTang13iccv}
LiMin Wang, Yu~Qiao, and Xiaoou Tang.
\newblock {Mining Motion Atoms and Phrases for Complex Action Recognition}.
\newblock In \emph{Proceedings of the IEEE International Conference on Computer
  Vision (ICCV)}, 2013{\natexlab{b}}.

\bibitem[Welinder et~al.(2010)Welinder, Branson, Mita, Wah, Schroff, Belongie,
  and Perona]{welinder10tr}
Peter Welinder, Steve Branson, Takeshi Mita, Catherine Wah, Florian Schroff,
  Serge Belongie, and Pietro Perona.
\newblock Caltech-ucsd birds 200.
\newblock Technical report, California Institute of Technology, 2010.

\bibitem[Yang et~al.(2011)Yang, Wang, and Mori]{yang11cvpr}
Weilong Yang, Yang Wang, and Greg Mori.
\newblock Recognizing human actions from still images with latent poses.
\newblock In \emph{Proceedings of the IEEE Conference on Computer Vision and
  Pattern Recognition (CVPR)}, 2011.

\bibitem[Yang and Ramanan(2011)]{confcvprYangR11}
Yi~Yang and Deva Ramanan.
\newblock Articulated pose estimation with flexible mixtures-of-parts.
\newblock In \emph{Proceedings of the IEEE Conference on Computer Vision and
  Pattern Recognition (CVPR)}. IEEE, 2011.

\bibitem[Yang and Ramanan(2013)]{yang12pami}
Yi~Yang and Deva Ramanan.
\newblock Articulated human detection with flexible mixtures of parts.
\newblock \emph{IEEE Transactions on Pattern Analysis and Machine Intelligence
  (TPAMI)}, 35, 2013.

\bibitem[Yao et~al.(2011{\natexlab{a}})Yao, Gall, Fanelli, and Gool]{yao11bmvc}
Angela Yao, Juergen Gall, Gabriele Fanelli, and Luc~Van Gool.
\newblock Does human action recognition benefit from pose estimation?
\newblock In \emph{Proceedings of the British Machine Vision Conference
  (BMVC)}, 2011{\natexlab{a}}.

\bibitem[Yao and Li(2012)]{yao12tpami}
Bangpeng Yao and Fei-Fei Li.
\newblock Recognizing human-object interactions in still images by modeling the
  mutual context of objects and human poses.
\newblock \emph{IEEE Transactions on Pattern Analysis and Machine Intelligence
  (TPAMI)}, 34\penalty0 (9):\penalty0 1691--1703, 2012.

\bibitem[Yao et~al.(2011{\natexlab{b}})Yao, Jiang, Khosla, Lin, Guibas, and
  Fei-Fei]{yao11iccv}
Bangpeng Yao, Xiaoye Jiang, Aditya Khosla, Andy~Lai Lin, Leonidas~J. Guibas,
  and Li~Fei-Fei.
\newblock Action recognition by learning bases of action attributes and parts.
\newblock In \emph{Proceedings of the IEEE International Conference on Computer
  Vision (ICCV)}, Barcelona, Spain, November 2011{\natexlab{b}}.

\bibitem[Yeffet and Wolf(2009)]{yeffet09iccv}
Lahav Yeffet and Lior Wolf.
\newblock Local trinary patterns for human action recognition.
\newblock In \emph{Proceedings of the IEEE International Conference on Computer
  Vision (ICCV)}, 29 2009-oct. 2 2009.

\bibitem[Yuan et~al.(2009)Yuan, Liu, and Wu]{yuan09cvpr}
Junsong Yuan, Zicheng Liu, and Ying Wu.
\newblock Discriminative subvolume search for efficient action detection.
\newblock In \emph{Proceedings of the IEEE Conference on Computer Vision and
  Pattern Recognition (CVPR)}, 2009.

\bibitem[Zhang et~al.(2011)Zhang, Khan, and Gotoh]{zhang11iccvw}
Lei Zhang, Muhammad Usman~Ghani Khan, and Yoshihiko Gotoh.
\newblock Video scene classification based on natural language description.
\newblock In \emph{Proceedings of the IEEE International Conference on Computer
  Vision Workshops (ICCV Workshops)}, pages 942--949. IEEE, 2011.

\bibitem[Zhou et~al.(2004)Zhou, Bousquet, Lal, {Jason Weston}, and
  Sch\"{o}lkopf]{Zhou2004}
Dengyong Zhou, Olivier Bousquet, Thomas~Navin Lal, {Jason Weston}, and Bernhard
  Sch\"{o}lkopf.
\newblock {Learning with Local and Global Consistency}.
\newblock In \emph{Advances in Neural Information Processing Systems (NIPS)},
  2004.

\bibitem[Zinnen et~al.(2009)Zinnen, Blanke, and Schiele]{zinnen09iswc}
Andreas Zinnen, Ulf Blanke, and Bernt Schiele.
\newblock An analysis of sensor-oriented vs. model-based activity recognition.
\newblock In \emph{ISWC}, 2009.

\end{thebibliography}
\end{document}